\documentclass[11pt]{article}
\usepackage{fontspec}
\usepackage{microtype}

\usepackage{amsmath,amssymb,amsthm}
\usepackage{graphicx}
\usepackage{booktabs}
\usepackage{authblk}
\usepackage{enumitem}

\usepackage{unicode-math}
\usepackage{polyglossia}
\setdefaultlanguage{english}
\setotherlanguage{arabic}
\newfontfamily\arabicfont[Script=Arabic]{Amiri-Regular.ttf}
\newfontfamily\devanagarifont[Script=Devanagari]{NewCM10Devanagari-Regular.otf}
\newcommand{\devazero}{{\devanagarifont ०}}

\newfontfamily\suppfont{FandolSong-Regular.otf}
\newcommand{\supp}[1]{{\suppfont #1}}

\usepackage[margin=1in]{geometry}
\usepackage[export]{adjustbox}

\usepackage[hyperfootnotes=false]{hyperref}
\hypersetup{
  colorlinks=true,
  linkcolor=BrickRed,
  citecolor=Green,
  filecolor=Mulberry,
  urlcolor=NavyBlue,
  menucolor=BrickRed,
  runcolor=Mulberry,
  pdfpagemode=FullScreen,
}
\usepackage[
  backend=biber,
  style=authoryear,
  citestyle=authoryear,
  giveninits=true,
  maxcitenames=3,
  maxbibnames=99,
  backref=true,
  dashed=false,
]{biblatex}
\renewbibmacro{in:}{}

\usepackage[title, titletoc]{appendix}

\usepackage[cachedir=_minted,frozencache]{minted}

\makeatletter
\let\c@table\c@figure
\let\c@listing\c@figure
\let\ftype@table\ftype@figure
\let\ftype@listing\ftype@figure
\makeatother

\usepackage{numprint}
\npthousandsep{,}
\usepackage[dvipsnames]{xcolor}
\usepackage[font=footnotesize]{caption}
\usepackage{wrapfig}
\usepackage{tablefootnote}

\usepackage{titlesec}
\titleformat{\section}{\normalfont\Large\bfseries}{\thesection}{0.7em}{}
\titleformat{\subsection}{\normalfont\large\bfseries}{\thesubsection}{0.7em}{}
\titleformat{\subsubsection}{\normalfont\large\bfseries}{\thesubsubsection}{0.7em}{}
\titlespacing*{\section}{0pt}{1ex}{0.1em}
\titlespacing*{\subsection}{0pt}{1ex}{0pt}
\titlespacing*{\subsubsection}{0pt}{1ex}{0pt}
\titlespacing*{\paragraph}{0pt}{1ex}{1em}

\title{\vspace{-1.0cm}\bfseries\Large
  Institutional Books --- Enriched Text \\
  {\large A customizable multilingual open-source pipeline for denoising, deduplicating, and annotating OCR text at scale}
}

\newcommand{\corrauthor}{\textsuperscript{$\dagger$}}

\theoremstyle{definition}

\newtheorem*{remark*}{Remark}

\newtheorem*{remarks*}{Remarks}

\newcommand{\ibet}{IB-HL-ET}
\newcommand{\centstack}[1]{%
  \begin{tabular}[c]{@{}c@{}}#1\end{tabular}%
}
\newcommand{\tallparens}[1]{%
  $\left(\vcenter{\hbox{#1}}\right)$%
}

\author[1]{David Lowry-Duda}
\author[1]{Matteo Cargnelutti}
\author[1]{Catherine Brobston}
\author[2]{Salwa Ismail}
\author[1]{Greg Leppert\corrauthor}
\author[3]{Amanda Watson}
\author[4]{Jonathan Zittrain}

\affil[1]{\small Institutional Data Initiative, Harvard Law School Library}
\affil[2]{\small Harvard Library}
\affil[3]{\small Harvard Law School Library}
\affil[4]{\small Harvard Law School, Harvard School of Engineering and Applied Sciences, Harvard Kennedy School}

\date{} 
\makeatletter
\renewcommand{\maketitle}{%
  \noindent{\small Technical report preprint.}\par
  \vspace{-1.5em}
  \noindent\rule{\textwidth}{2pt}
  \vspace{1.5em}

  \begin{center}
    {\bfseries\Large \@title \par}
    \vspace{1.1em}
    \rule{\textwidth}{2pt}
    \vspace{0.8em}

    {\normalsize
      \@author
    \par}
  \end{center}
  \vspace{1.0em}
}
\makeatother

\renewenvironment{abstract}{
  \begin{center}
    \bfseries Abstract
  \end{center}
  \vspace{-0.3em}
  \small
}{\par\vspace{1.2em}\normalsize}

\begin{document}
\maketitle

\begingroup
\renewcommand\thefootnote{$\dagger$}
\footnotetext{Corresponding author \href{mailto:gleppert@law.harvard.edu}{gleppert@law.harvard.edu}}
\addtocounter{footnote}{1}%
\endgroup

\vspace{-1em}
\begin{abstract}

Released in 2025, Institutional Books: Harvard Library (IB-HL) is a collection
of 983,004 volumes (242B \textsc{o200k\_base} tokens),
originally digitized through Harvard Library's participation in the Google Books
Library project. As
researchers and developers have begun to use IB-HL, a tension has emerged between standard
large-scale preprocessing practices and the goals of careful information stewardship. Many existing
pipelines optimize for web text: as a result, they tend to aggressively filter, deduplicate, restrict by language, and
sometimes discard meaningful metadata. Meanwhile, researchers seeking to use IB-HL duplicate effort while performing
similar processing and analysis.

We describe an approach that we call \textbf{Enriched Text}. Instead of producing a single
``complete'' stream of tokens, we normalize the text while preserving metadata through annotations.
We separate endmatter, detect per-paragraph language, identify clusters of duplicate paragraphs, and
compute per-paragraph bits-per-byte scores. We provide this information through HTML-like
annotations layered on top of the text. By parsing these annotations, users can tailor the output to
their own needs  instead of accepting a global editorial decision on content.
The pipeline applies to all $\approx$250 languages in the collection.

This report describes this project's goals, implementation, and design rationale.
The release includes \ibet{} (an enriched-text version of IB-HL containing 217B
\textsc{o200k\_base} tokens across \numprint{983003} volumes, organized into 1.39B annotated
subtopic paragraphs) and the pipeline that produced it. These serve to make the
collection easier for machines to parse and for humans to study.

\end{abstract}

\clearpage{}

{
\setlength{\parskip}{0.15em}
\renewcommand{\baselinestretch}{0.30}\normalsize
{\small
\tableofcontents{}
}
}

\clearpage{}

\section{Introduction}
\setcounter{footnote}{0}

In 2025, the Institutional Data Initiative introduced Institutional Books: Harvard Library (IB-HL)\footnote{\url{https://huggingface.co/datasets/institutional/institutional-books-1.0}}~\parencite{cargnelutti2025institutional},
a 242B-token dataset of digitized books and other bound materials drawn from Harvard
Library's collections.
That work emphasized rigorous information stewardship, including detailed provenance,
collection-level deduplication, optical character recognition (OCR) artifact analysis,
post-processing to improve legibility, and rich metadata.

Following the release of IB-HL, researchers and developers sought to use the corpus for large
language model (LLM) training and processing.
\textsc{TypewriterLM}\footnote{\url{https://huggingface.co/typewriter-ai/typewriter-1913-7B-base}}~\parencite{luo2026pretraining}
uses IB-HL as its primary source (97.7\% of tokens) for its 54B-token historical
pretraining corpus, for a 7.2B-parameter language model trained on pre-1913 English.
Similarly, the
\textsc{talkie-1930}\footnote{\url{https://huggingface.co/talkie-lm/talkie-1930-13b-base}}~\parencite{levine2026talkie}
13B-parameter ``vintage language model'' uses IB-HL data and metadata in its 260B-token pre-1931
English corpus.
The \textsc{K2-V2} open-weight reasoning model~\parencite{liu2025k2}
incorporates IB-HL as a source of long-context documents in its mid-training data.

Feedback from early adopters indicates tensions between common large-scale
preprocessing practices and the goals that motivated IB-HL\@.
LLMs depend on the nature, quality, and diversity of the data used in
training, yet standard internet-oriented pipelines often rely on aggressive language
filtering (such as only using English) and deduplication (such as discarding all near-duplicate
text); see for example~\parencite{kaplan2020scaling, gunasekar2023textbooks,
muennighoff2023scaling}. These practices strip multilingual content and reduce document fidelity.
For IB-HL, excluding non-English would discard half the corpus and narrow its
cultural scope.
Naive chunk-level deduplication fragments books and long-form documents.
Furthermore, pipelines often flatten or discard \emph{paratext}~\parencite{genette91paratext}: titles,
tables of contents, footnotes, indices, publisher notices, and other elements that mediate between a
text and its readers.

We describe an alternative approach that we refer to as \textbf{Enriched Text}.
Rather than producing a single ``clean'' stream of tokens, we offer a configurable dataset.
Concretely, we represent many paratextual elements and metadata (e.g.\ front and back matter,
detected languages, duplicate paragraph information) as machine-readable annotations layered on top
of the OCRed text.
We also try to identify and separate the \emph{endmatter} (consisting of the front and back matter
framing the main text, including the title page, copyright page, dedication, table of contents,
foreword, glossary, bibliography, index, author's notes) from the \emph{middlematter} (i.e.\ the
main text itself).
These annotations are made available as attributes in an HTML-like markup language that can be
quickly and efficiently parsed with standard parsing libraries, ultimately enabling users to tailor
the representation to their needs.

This work focuses on three foundational problems in large-scale dataset construction:
\begin{enumerate}[nosep,noitemsep]
  \item \textbf{Cleaning} OCRed text across hundreds of languages and scripts while preserving
    underlying structure;
  \item \textbf{Segmenting} volumes into semantically meaningful chunks that respect document and section boundaries; and
  \item \textbf{Identifying Duplicate Segments} across the whole collection.
\end{enumerate}
By releasing \ibet{}, we continue IDI's broader project of establishing a
community-led, knowledge-institution-anchored approach to data.
We offer a variant of IB-HL that integrates into modern
information-access workflows, training pipelines, and other forms of computational access and scholarship.

\section{Contributions}

We introduce the following contributions:

\begin{enumerate}[nosep,noitemsep]
  \item
    A detailed technical report on the processing work we have conducted to
    create \ibet{}, an \textbf{Enriched Text} version of the Institutional Books
    --- Harvard Library dataset sourced from Harvard Library's collection.

  \item
    A public dataset containing 217B \textsc{o200k\_base} tokens across
    7B sentences, 1.39B subtopic paragraphs, 297M subtopic sections, and 983k
    volumes.
    \\
    Available at
    \url{https://huggingface.co/datasets/institutional/institutional-books-hl-enriched-text}.

  \item
    The Python pipeline we created to produce this collection.
    \\
    Available at \url{https://github.com/institutional/institutional-books-enriched-text-pipeline}.

  \item
    A small, dependency-free Python library for parsing and adapting the dataset to
    specific use cases.
    Available at
    \url{https://github.com/institutional/institutional-books-enriched-text-parser}.

  \item
    Synthetically generated training data for an endmatter classifier and subclassifier.
    Available at \url{https://github.com/institutional/institutional-books-enriched-text-pipeline}
    as a GitHub release.
\end{enumerate}

\section{Goals and Scope}

\ibet{} is another step in a multifaceted, collaborative research process. Our
goal is to make the underlying collection easier to access \emph{computationally}, i.e.\
easier to read, filter, search, and study using computational methods, including
AI\@. We treat this as a text-to-text problem, extending previous OCR and
OCR-refining work. We organize and analyze existing text, but we do not re-OCR
any volume. This allows the pipeline to be applied downstream of, and
efficiently augment, more computationally expensive processes the outputs of
which organizations may not want to discard or recreate. This may be especially
true where organizations inherit existing OCR through vendor partnerships (e.g.\
Google Books) or their own digital archives.

\paragraph{Historical Collection.}
This project extends IB-HL and the source material from the Harvard Library collection.
The same disclaimers from IB-HL apply; we advise readers to consult the
disclaimer in \S\ref{sec:disclaimers}.

\paragraph{Configurability.}
To improve computational access, we separate the linguistic content of a volume (its words and
punctuation) from its \emph{paratext} (reference materials that surround and mediate the main text).
We prefer to label this material rather than delete it. In particular, we identify and separate
endmatter from middlematter so that users can distinguish the \emph{text} from its \emph{context},
but still draw on that context when useful. More generally, we annotate the text
so that users can \emph{choose} how to interact with the collection. A reader
who wants long-context English prose can strip all endmatter and filter only to English text using
only annotations, while
readers with different use cases can retain the layers and data they need.

In keeping with the stewardship goals of IB-HL, we
prioritize fidelity to the source material and prefer to \emph{mark} and \emph{annotate}
rather than rewrite or remove.
However, in building a highly configurable dataset and pipeline, we made
some decisions that may marginally impact fidelity.
We believe these greatly improve usability; unmodified OCR remains available in
IB-HL for users with different requirements.

Specifically, we
remove page numbers and running headers and footers (cf.\ \S\ref{ssec:rhrf},
\S\ref{ssec:pagenums}).
These elements are marginalia and have different organizational granularity than
the sequence of sentences and paragraphs in the main text. Though marginalia
carries useful navigational and bibliographic value, once page boundaries are
removed it appears mostly as recurring noise and fragments for the purposes of
large-scale computational access. Addressing these forms of paratext (and
marginalia more broadly) is left to future work.

\paragraph{Available Compute.}
The techniques described here are designed to be widely available and reproducible.
For both responsible use of resources and reproducibility, we favor frugal
computation~\parencite{vanderbauwhede2023frugal}, echoing a philosophy of IB-HL. Most of the
pipeline runs on CPUs or involves small machine learning (ML) models that run on a single GPU, and we set an
approximate target that the whole collection be reproducible within one month on a single NVIDIA DGX
Spark.\footnote{This pipeline hits this target with one major exception: computing bits-per-byte for
  all paragraphs. Though we chose a small 0.6B model for this computation, this step
requires months of dedicated GPU time. See \S\ref{ssec:bpb} for details.}

To run the pipeline, we used the Harvard FASRC Cannon cluster.
Knowing we would run this pipeline on a cluster, we designed the pipeline around a
distributed computing workflow with systematic error detection and resolution. Ultimately, the
\emph{library} powering the pipeline is efficient and modest in its requirements, while the
\emph{orchestration} optimizes for large, heterogeneous compute availability.

Throughout, we report the actual compute times required on our cluster or, when relevant, on a
single machine. The Harvard Cannon cluster has heterogeneous hardware.
During execution, the typical compute node had at least 4 Intel Cascade Lake CPU cores; the typical
GPU computation node had at least 4 Intel Sapphire Rapids CPU cores and one NVIDIA A100 80GB GPU\@.
We aim to provide enough information to allow others to assess the cost of reproducing or extending
each stage of computation.

\clearpage{}

\section{Processing Pipeline}%
\label{sec:pipeline}

\begin{wrapfigure}{R}{0.50\textwidth}
  \centering
  \includegraphics[width=0.98\linewidth]{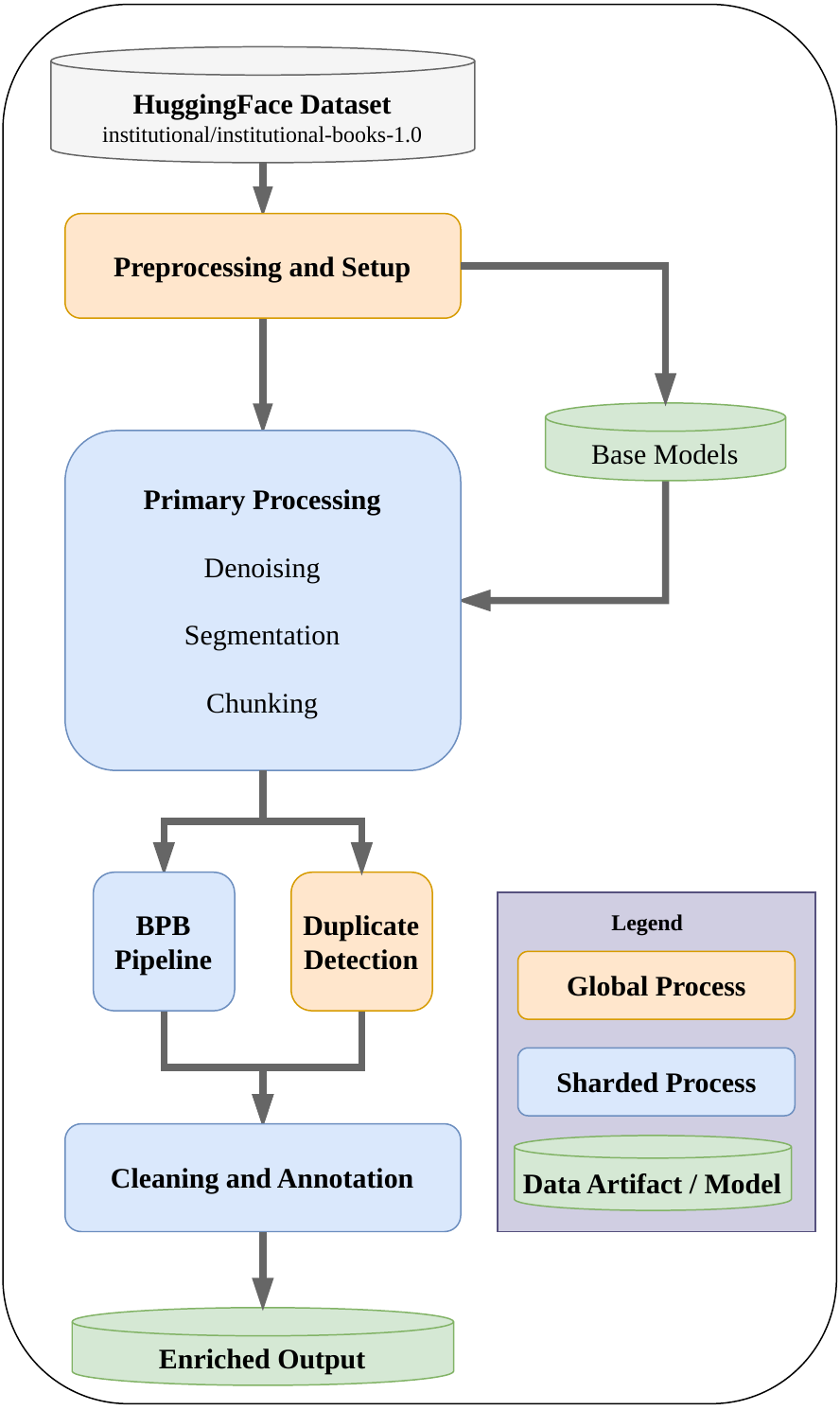}    
  \caption{Pipeline Overview}\label{fig:pipeline_overview}
  \vspace*{-4em}
\end{wrapfigure}

See Figure~\ref{fig:pipeline_overview} for an overview of the processing pipeline.
The input is the IB-HL dataset,
consisting of 242B \textsc{o200k\_base} tokens across \numprint{983004} volumes.
The pipeline is a sequence of text-to-text transformations.
In this section, we describe the aims and methodology of each step; several steps have additional
details in Appendix~\ref{appendix:details}.

The first step is preprocessing and setup (cf.\ \S\ref{ssec:preprocessing}).
This step also splits the dataset into shards.
Most remaining steps share no memory or information across shards and are suited for parallel
computation.
Each shard goes through Unicode normalization (cf.\ \S\ref{ssec:unicode}), duplicate page removal
(cf.\ \S\ref{ssec:duplicate_page}), endmatter separation (cf.\ \S\ref{ssec:endmatter}),
dehyphenation (cf.\ \S\ref{ssec:dehyphenation}), running header and footer
removal (cf.\ \S\ref{ssec:rhrf}), page number removal (cf.\ \S\ref{ssec:pagenums}), sentence
segmentation (cf.\ \S\ref{ssec:segment}), and subtopic section chunking  (cf.\ \S\ref{ssec:chunk}).

By this stage, each volume's text is organized in subtopic paragraph chunks.
We identify (but do not remove) duplicate paragraphs across the whole collection (cf.\
\S\ref{ssec:deduplication}) and score every paragraph by its bits-per-byte (BPB) (cf.\
\S\ref{ssec:bpb}). Finally, we detect the language of each paragraph, assemble
the metadata, and format the output (cf.\ \S\ref{ssec:annotation}).

\subsection{Preprocessing}\label{ssec:preprocessing}

Later stages in the pipeline apply techniques informed by the output of different base language
models.
During preprocessing, the pipeline trains the base language models and assembles the data into
shards.
The pipeline also distills a static multilingual embedding model.

\subsubsection{Methodology}\label{sssec:preprocessing_method}

\paragraph{Shard Assembly.} We group the \numprint{983004} volumes of IB-HL into shards of up to
200 volumes each. Books are grouped into two categories depending on the
primary volume language (as determined by IB-HL).
One category consists of volumes whose primary language is one of the $138$ languages listed in
Table~\ref{table:nupunkt_languages} in Appendix~\ref{appendix:ss:segmentation}; we later segment
these texts into sentences (cf.\ \S\ref{ssec:segment}) using the \textsc{Nupunkt}
library~\parencite{nupunkt_paper} and refer to these as ``\textsc{Nupunkt}-compatible'' volumes.
The other category consists of all remaining languages.
Each shard is assembled from volumes in exactly one category.

This distinction optimizes for sentence segmentation.
\textsc{Nupunkt}-compatible shards can be processed (up to BPB annotation) without GPUs, whereas the
other shards require GPUs for efficient sentence segmentation.

\paragraph{Base Language Models.} Base language models are bootstrapped from the corpus itself
for improved handling of historical language patterns.
We select up to $30$ volumes from each language (as identified per volume in IB-HL) when available.
Within each language subcollection, we gather n-gram statistics ($1$-grams to $5$-grams) for a small
n-gram language model.

N-gram language models are trained on a heavily normalized version of the text: full NFKC
Unicode normalization with additional normalizations applied to hyphens, spaces, quotes, and whitespace
(cf.\ Appendix~\ref{appendix:sss:unicode} ``hard normalization''). These models are used in
dehyphenation (cf.\ \S\ref{ssec:dehyphenation}).

For each of the $138$ \textsc{Nupunkt}-compatible languages, we treat each subcollection
as a training corpus for a base \textsc{Nupunkt} model.
Each \textsc{Nupunkt} corpus undergoes the same Unicode normalization described in \S\ref{ssec:unicode}
to match the normalization during inference.

\paragraph{Static Embedding Models.}
We use \textsc{Model2Vec}~\parencite{model2vec} to distill \textsc{BAAI/BGE-M3}~\parencite{bgem3}
into a static embedding model.
This distilled embedding model is later used in chunking (cf.\ \S\ref{ssec:chunk}).
We choose \textsc{BAAI/BGE-M3} due to its multilingual support.
Distilling \textsc{BAAI/BGE-M3} to a static model using \textsc{Model2Vec} allows
inference to be performed using only CPUs. 
Though this lowers precision and accuracy slightly, the gain in computational efficiency enables us
to use embeddings throughout the pipeline.

Further, we fine-tune the static embedding model into a static classifier and static subclassifier
used for endmatter classification (cf.\ \S\ref{ssec:endmatter}).
The data used for fine-tuning was synthetically generated and is distributed as GitHub release
assets (cf.\ \S\ref{sssec:endmatter_methods}).
We separated the training data into an 80-20 split and oversampled classes  
during training to balance categories.

\subsubsection{Results}\label{sssec:preprocessing_results}

Preprocessing lasted 9.5h on 4 cores of a shared Intel Cascade Lake node on a compute cluster.
\textsc{Nupunkt} model training required 3h (averaging $\approx78.4$ s/language). N-gram
training required 2.2h (averaging $\approx31.5$ s/language). Distillation required
1h, and training the classifier and subclassifier required 0.5h each. Data transfer occupied the
remaining time.

In total, $388$ base models and two endmatter classification models were trained during preprocessing:
\begin{enumerate}[nosep,noitemsep]
  \item Base n-gram language models (in 250 languages) were trained on subcollections of up to 30
    books in each language.
  \item Base \textsc{Nupunkt} segmentation models were trained on subcollections of up to 30 books in each of
    the 138 \textsc{Nupunkt}-compatible languages.
  \item One endmatter classification model and subclassification model were trained.
\end{enumerate}
Only 82 languages had the full $30$ books available in the corpus for base model training.
The remaining 168 languages have ``base'' models built from every book in that language in IB-HL.

The trained endmatter classifier detects endmatter with $0.97$ accuracy (F1 $0.97$). The trained
subclassifier is less accurate, with $0.90$ overall accuracy.

\subsection{Unicode Normalization}\label{ssec:unicode}

Overcoming the challenges posed by OCR reliability in historical document digitization is an active
area of research.
Unicode normalization is one basic step towards addressing artifacts and differences in
historical typography~\parencite{neudecker2021survey, beyene2026survey}.

IB-HL contains volumes spanning centuries of typography.
The same underlying text can appear in different byte-level forms.
For example, the accented character (\'{e}) could be encoded as U+00E9 (\'{e}) or U+0065 and
U+0301 (e and {} ́), depending on OCR preference and configuration.
OCR treatment is inconsistent across ligatures, diacritics, full-width and superscript text,
whitespace, zero-width characters, hyphen and dash code points, curly and straight quotation marks,
and scientific or mathematical notation.
We follow standard Unicode normalization practices~\parencite{unicode_tr15, unicode_icu} to
facilitate text mining and data preparation.

We choose to prioritize accountability and provenance of the underlying source material:
we conservatively normalize text, but do not seek to detect and correct individual OCR
mistakes.

\subsubsection{Methodology}\label{sssec:unicode_method}

We have two normalization regimes with different aims: \emph{soft} normalization and \emph{hard}
normalization.
The exact transformations are in Listing~\ref{listing:unicode} in
Appendix~\ref{appendix:sss:unicode}.

Soft normalization is conservative. We first apply Unicode NFC composition. The Unicode standard specifies
the U+200B zero-width space as a formatting hint for word break and line break opportunities in
languages that have no visible word spacing, such as Thai, Myanmar, Khmer, and
Japanese (see \S23.2 ``Layout Controls'' in \cite{unicode_standard_v15}). OCR engines sometimes output
zero-width spaces, but these are inconsistent and manifest as invisible noise in textual analysis. We
follow the common practice of
removing U+200B zero-width spaces. We explicitly preserve \emph{visible} text, including line
breaks, curly quotes, hyphens, dashes, ligatures, and accents. Each step resolves encoding-level
inconsistencies without altering how the text visibly reads.
All output text in \ibet{} is soft-normalized.

Hard normalization is lossy and only used internally to determine when two different
passages refer to the same reduced text. In addition to the soft normalization's operations, hard
normalization applies NFKC compatibility normalization, maps space-like characters to ASCII space,
maps every quote to ASCII quotes, and flattens all tabs and newlines. The output is a single line
with ASCII punctuation and NFKC-folded characters.
Hard normalization is never stored and never presented to users.
It is computed transiently when robustness against typographic variation matters more
than fidelity, namely for dehyphenation (cf.\ \S\ref{ssec:dehyphenation}), exact duplicate page removal
(cf.\ \S\ref{ssec:duplicate_page}), and duplicate passage identification (cf.\
\S\ref{ssec:deduplication}).

\subsubsection{Results}\label{sssec:unicode_results}

A total of 46h (combined wall-time across 1-core jobs on Intel Cascade Lake workstations) was spent on soft-normalizing
all the text at the start of the pipeline. Per-volume cost was uniform and inexpensive ($\approx 0.17$ s/volume).

\subsection{Duplicate Page Removal}\label{ssec:duplicate_page}

Within each volume we detect and discard duplicate pages.
The vast majority of these examples are repeated scans from the initial scanning process, though
some recurring text is also removed.
This is distinct from identifying duplicate paragraphs (cf.\ \S\ref{ssec:deduplication}).

\subsubsection{Methodology}

Scanning the same physical page twice usually yields slightly different OCRed text.
Thus we search for \emph{near}-duplicate pages.

Each page is assigned a 128-bit simhash~\parencite{charikar2002simhash}
assembled from 9-grams.
The simhash is assembled using the MurmurHash3\footnote{MurmurHash3 is optimized for efficient
  computation. We capture more of this efficiency by writing a small C++ extension that computes and
compares simhashes with MurmurHash3, avoiding the overhead of passing Python objects back and forth.
See Appendix~\ref{appendix:ss:duplicate_page}.}
algorithm (based closely on the original implementation~\parencite{smhasher}).
Pages with fewer than $50$ characters are not candidates for duplicate removal
to prevent spurious removal of short pages.
Initial tests showed that repetitive low-entropy text could dominate the final
simhash.
For example, pages with the text ``\texttt{............}''  
have biased simhashes, but are common in tables and lists.
To mitigate this problem, we only hash 9-grams with at least 4 distinct characters.

Two simhashes are considered near-duplicate if they differ by at most 6 bits.
We compare all qualifying pages' simhashes pairwise within each volume. We group mutually
near-duplicate pages with a union-find (as in Chapter~21 of the standard algorithms book
by~\cite{cormen2009introduction}). Thus if page $A$ is a near-duplicate of page $B$ and page $B$ is
a near-duplicate of page $C$, then page $A$ is noted as a near-duplicate of page $C$ even if their
simhashes differ by more than $6$ bits.
Choosing a $6$ bit threshold is conservative and gives a low chance of spurious duplicate removal
even after union-find accumulation.

\subsubsection{Results}\label{sssec:duplicate_page_results}

Page-level deduplication required a total of 164h (combined wall-time across 1-core jobs on Intel
Cascade Lake workstations), with approximately equal amounts of time spent on (1) transient hard
Unicode normalization, (2) simhash computation, and (3) pairwise simhash comparison. We show a few
statistics at the top of Table~\ref{table:duplicate_page_rates}.

\begin{table}[htb!]
  \centering
  {\small
    \begin{tabular}{lc}
      \toprule
      Books Processed\footnotemark{} & \numprint{983003} \\
      Total duplicate pages removed & \numprint{1981296} \\
      Books with at least 1 duplicate page & \numprint{180521} \\
      Mean pages removed per affected volume & 10.98 \\
      \bottomrule
    \end{tabular}
  }

  \vspace*{1em}

  \centering
  {\small
    \begin{tabular}{llccc}
      \textbf{Barcode}& \textbf{Lang}& \textbf{Pages Removed}& \textbf{Total Pages}& \textbf{\% of Book Removed} \\
      \midrule
      32044081823502& English&  560&           692   &           80.9\%  \\
      32044097046809& English&  338&           464   &           72.8\%  \\
      32044011481660& English&  894&           1382 &           64.7\%  \\
      32044050769256& French&  920&           1448 &           63.5\%  \\
      32044004373478& English&  46 &           76    &           60.5\%  \\
      32044018915694& English&  339&           566   &           59.9\%  \\
      32044011481652& English&  721&           1238 &           58.2\%  \\
      32044088907993& English&  720&           1256 &           57.3\%  \\
      32044085160604& Latin&  106&           188   &           56.4\%  \\
      HL01BW        & English&  499&           886   &           56.3\%
    \end{tabular}
}\caption{Books with High Duplicate Page Rates}\label{table:duplicate_page_rates}
\end{table}

Approximately 81.6\% of books had no duplicate pages.
Some books had a substantial portion of duplicate pages.
The 10 most extreme are indicated at the bottom of Table~\ref{table:duplicate_page_rates}.
These examples have unusually repetitive page scans. The book with barcode
32044081823502 (titled \emph{Memoir of Isaac Richardson}) has 132 pages post-deduplication, while
containing 692 page scans (of which only 71 are middlematter).

\subsection{Endmatter Separation}\label{ssec:endmatter}

A trained page classifier splits each volume into frontmatter, middlematter, and
backmatter.
We output the separated endmatter largely unprocessed.
A trained subclassifier categorizes each endmatter page into one of
\texttt{TOC\_INDEX} (table of contents or index pages), \texttt{BIBLIO} (bibliography, citations, or
lists of references), or \texttt{OTHERENDMATTER}.
The remaining pipeline operates primarily on middlematter.

\subsubsection{Methodology}\label{sssec:endmatter_methods}

\footnotetext{There is exactly one book in IB-HL that is not in \ibet{}. It has the title
  \emph{Nederlandsche pasicrisie: bevattende in alphabetische methode den zakelijken inhoud van alle
    in Nederland gewezen regterlijke beslissingen en de chronologische listj dier beslissingen, met
    opgave van al de verzamelingen van regtspraak waarin zij gevonden worden: alphabetisch gedeelte
    Chrono.\ gedeelte: 1887}, barcode \textsc{HL0ONK}. This book is omitted because the entire book
  is one large table with correspondingly poor OCR\@. The endmatter classifier flagged this book as
exceptional as almost the entire book was detected as endmatter. }
Reliably detecting endmatter across a diverse collection is known to be challenging.
Our approach is modeled on the experimental OCR text post-processing strategy from
IB-HL (see \S4.9 of~\cite{cargnelutti2025institutional}).
We use the \textsc{Model2Vec}-distilled \textsc{BAAI/BGE-M3} static embedding model described in
\S\ref{sssec:preprocessing_method} and its static fine-tunes for the classifier and subclassifier.

\paragraph{Synthetic Data.}
This pipeline uses synthetic data for fine-tuning.
On one hand, testing indicates that large-scale annotation of actual endmatter leads to stronger models.
On the other hand, testing also indicates that multilingual support is essential for training a
multilingual classifier. We deliberately preserve the multilingual nature of the collection,
but could not collect high-quality annotations across such a diverse set of languages.
Recent research has shown that synthetically generated training data is
effective at improving model performance when limited training data is
available~\parencite{xie2020unsupervised, gunasekar2023textbooks, smollm}. Further, performance is better for simple,
non-subjective tasks after careful curation~\parencite{li2023synthetic, li2024data}.
Thus using multilingual synthetic data is a pragmatic choice.

To generate synthetic data, we compared the behavior among
\textsc{gpt-oss-20b}~\parencite{openai2025gptoss120bgptoss20bmodel},
\textsc{gemma3}~\parencite{gemma},
and \textsc{qwen3}~\parencite{qwen3technicalreport}.
After creating and verifying samples in each of the models, we ultimately choose
to focus on \textsc{gpt-oss-20b} for its performance and permissive licensing.
See Listing~\ref{listing:synthetic} in the Appendix for the structure of our generation prompts.

We generated 150k training data examples. From these, we selected
\numprint{44532} valid training examples for endmatter-vs-middlematter and \numprint{27192} examples
for classifying endmatter.
Curation involved several rounds of cleaning and pruning.
Generated examples often included markdown; mid-resource languages produced many duplicate
examples; and low-resource languages led to vacuous outputs. We stripped markdown, removed
duplicates, enforced a minimum length requirement, and checked claimed languages using
\textsc{pyfranc}~\parencite{pyfranc}.

\paragraph{Inference.}
We treat endmatter separation conservatively and non-destructively. Each book's detected frontmatter
consists of all pages until the first non-endmatter page, and similarly (in reverse) for the
backmatter. Pages with only whitespace are not counted as middlematter when
determining endmatter boundaries.

The pipeline classifies all pages in each book in batches of up to $1024$ pages at a
time. This leads to more middlematter classifications than necessary, but
most books are classified in a single batch and this leads to little overhead.

\paragraph{Subclassification.}
We choose to focus on the subtypes \texttt{TOC\_INDEX} (Table of Contents), \texttt{BIBLIO}
(Bibliography), and \texttt{OTHERENDMATTER}. Both tables of contents and indices tend to have
similar grammars and typesetting, making them difficult to distinguish.
Bibliographies and lists of references are also easily confused. Other common endmatter pages
(dedication pages, publication pages, title pages, copyright pages, bibliographic information,
library or collection plates, and so on) vary in form and would require extensive annotation support
for accurate prediction.

\subsubsection{Results}\label{sssec:endmatter_results}

Endmatter classification lasted 211h (combined wall-time across many 4-core jobs on Intel Cascade
Lake workstations).
No significant timing differences occurred across languages. This is only computationally viable
due to the \textsc{Model2Vec} static model architecture.

On average, each book has approximately $14$ pages of endmatter.
Endmatter pages have additional sources of OCR errors (e.g.\ hand-written additions, library
stamps, book plates), leading to higher variance in classification.
The resulting errors more often push endmatter into middlematter than the reverse.
For example, one common overly-conservative signal to start endmatter is when a book's call number
is handwritten on some frontmatter page, but OCRed incorrectly.

\begin{table}[htb!]
  \centering
  {\small
    \begin{tabular}{llc}
      & \textbf{Pages} & \textbf{\% of all pages} \\
      \midrule
      Frontmatter & \numprint{7165654} & 1.86\% \\
      Backmatter & \numprint{6560012} & 1.70\% \\
      All Endmatter & \numprint{13725666} & 3.55\% \\
      Source Pages & \numprint{386256945} &  --- \\
    \end{tabular}

    \vspace{1em}

    \begin{tabular}{llllc}
      & \textbf{Frontmatter} & \textbf{Backmatter} & \textbf{Total Endmatter} & {\bf\shortstack{ \%
      Non-blank \\ Endmatter Pages}} \\
      \midrule
      \texttt{OTHERENDMATTER} &  \numprint{3163667}  &   \numprint{1232023} &   \numprint{4395690}  &  54.7\% \\
      \texttt{TOC\_INDEX}      &  \numprint{1294656}  &   \numprint{1639797} &   \numprint{2934453}  &  36.5\% \\
      \texttt{BIBLIO}         &  \numprint{156219}   &    \numprint{556543} &     \numprint{712762} &   8.9\% \\
      \midrule
      Total          &  \numprint{4614542}  &   \numprint{3428363} &   \numprint{8042905}  &  100\% \\
      \midrule
      Blank Pages & \numprint{2551112} & \numprint{3131649} & \numprint{5682761} & ---
    \end{tabular}
    \caption{Endmatter Classification Statistics}\label{table:endmatter_classification}
  }
\end{table}

Classification statistics are in Table~\ref{table:endmatter_classification}. Note that the ``Total''
line does not include blank pages.
About 41\% of frontmatter and backmatter pages are blank.
Backmatter has a higher blank-page rate than frontmatter (47.7\% vs.\ 35.6\%), possibly due to fixed
folio lengths and trailing sheets.

\subsection{Dehyphenation}\label{ssec:dehyphenation}
End-of-line (EOL) hyphens are often introduced to create more consistency in the physical appearance of
blocks of text.
Removed from their original context, these hyphens are artifacts
that can corrupt underlying words (e.g.\ writing \texttt{informa- / tion} instead of
\texttt{information}).

One strategy would be to detect and remove all EOL hyphens using regular
expressions. Applying this to the whole dataset, however, would collapse many
legitimate words and compound phrases into non-words. We instead resolve
end-of-line hyphens using an n-gram language model that ranks possible changes.

\subsubsection{Methodology}

For every line that ends with a hyphen/dash, we rank three options and choose the most likely:
\begin{enumerate}[nosep,noitemsep]
  \item merge the fragments into one (e.g.\ \texttt{informa- / tion} becomes \texttt{information}),
  \item remove the linebreak but keep the hyphen (e.g.\ \texttt{mother-in- / law} becomes\\
    \texttt{mother-in-law}), and
  \item replace the linebreak with a space and keep the hyphen (e.g., with interruptions or emdashes
    that happen to occur at ends of lines).
\end{enumerate}

Any given word is more likely to appear mid-text than only at line breaks.
A probability model trained on the text will capture these likelihoods, and we choose
n-gram statistics as our basic model.
Ranking is based on character n-grams (using $1$-grams to $5$-grams). We combine the base n-gram
model (trained during preprocessing, cf.\ \S\ref{ssec:preprocessing}) for the book's primary
language\footnote{As detected by IB-HL. If the language is ``unknown'', we default to English, 
the most common language in IB-HL.} with a book-specific n-gram model built transiently in
memory from the volume's own text.
We only train this book-specific model if the book has end-of-line hyphenation.
The base model includes per-language statistical information while the book-specific model includes
proper nouns, technical terms, and other book-specific vocabulary.

We estimate the probability of a specific n-gram $w_1^n := w_1 \cdots w_n$ by its add-$k$ smoothed
density
\[ P(w_1^n) \approx \frac{(\# w_1^n) + k}{(\# \text{n-grams seen}) + k \cdot (\text{vocab size})} \]
with $k = 0.001$.
This approach is modeled after standard n-gram language models (see for instance Chapter 3
of~\cite{jm3}), but with some simplifying assumptions for speed.
The add-$k$ smoothing is a crude form of Laplace smoothing and avoids exact $0$.
To determine the relative ranking among the three possibilities, we approximate the joint
probabilities of each option.
We estimate the joint probabilities of each possibility using a $10$-character
window on each side of the hyphen.

When an n-gram $w_1^n$ has never been seen, we approximate the probability using
the so-called \emph{Stupid Backoff}~\parencite{brants2007large} algorithm with backoff probability
of $0.4$ times the $(n-1)$-gram formed when omitting the first character,
\[ P(\text{unseen } w_1^n) \approx 0.4 \cdot P(w_2^n). \]
The resulting language model is not a true probability distribution, but it adequately ranks the
three possibilities and is faster than more sophisticated models.

Overall, the pipeline computes these two sets of probabilities using both the book-specific model
and the base language models, computes associated perplexities, and weighs both equally for the
final determination.
(See Appendix~\ref{appendix:ss:dehyphenation} for discussion of multiple other technical details and
alternatives considered. See Appendix~\ref{appendix:s:bad} for an example where dehyphenation
performs poorly on technical text).

\subsubsection{Results}

\begin{table}[htb!]
  \centering
  \begin{minipage}{0.50\linewidth}
  \centering
  {\small
    \begin{tabular}{lc}
      \toprule
      Total Hyphens Removed & \numprint{2322246287} \\
      Estimated EOL Hyphen Removal Rate & $\approx$ 89\% \\
      Estimated EOL Hyphen Keep Rate & $\approx$ 7\% \\
      Estimated EOL Hyphen+Space Rate & $\approx$ 4\% \\
      \bottomrule
    \end{tabular}
  }
  \end{minipage}
  \hfill
  \begin{minipage}{0.45\linewidth}
  \centering
  {\small
    \begin{tabular}{lc}
      \multicolumn{2}{c}{\footnotesize High Dehyphenation Languages} \\
      \midrule
      \textbf{Language} & \textbf{Avg Removed/Page} \\
      \midrule
      Latin & 11.3 \\
      Russian & 11.1 \\
      Bulgarian & 11.1 \\
      Polish & 10.3 \\
      Hungarian & 9.3 
    \end{tabular}
  }
  \end{minipage}
  \caption{Hyphenation Statistics}\label{table:dehyphenation}
\end{table}

A total of $640$h (combined wall-time across many 4-core jobs on Intel Cascade Lake
workstations) was spent on dehyphenation. Run statistics are in Table~\ref{table:dehyphenation}.
Typically (89\% of the time) the dehyphenation strategy removed the hyphen and newline.
This supports the widespread practice of using regular expressions to detect and
remove all end-of-line hyphens, but also suggests that this practice is not always appropriate.

Cyrillic, Latin, and European languages have by far the most inserted end-of-line hyphens.
In this collection, Latin has the highest per-page rate, consistent with dense scholarly printing of
classical text.
For English books, the average number of hyphens merged per page was $4.9$, substantially smaller
than Latin or the Slavic languages.
As expected, many languages have almost no end-of-line hyphens.
Chinese, Japanese, and Arabic all average below $0.1$ removed hyphens per page (and those with
hyphens are often multilingual).

\subsection{Running Header and Footer Removal}\label{ssec:rhrf}

Printed books often show a running header and/or footer containing the title, author, or chapter
name. OCR tends to place these among the first or last lines on a page. Removing page boundaries
causes these to become inline noise, disrupting paragraphs that span multiple pages. We identify
recurring short lines near the top or bottom of nearby pages and remove
them.\footnote{This removal strategy is not perfect. See Appendix~\ref{appendix:s:bad} for a
concrete example where this strategy removes too much text. Improved marginalia handling
is left to future work.}

\subsubsection{Methodology}

We identify running headers and footers by detecting locally concentrated clusters.
We allow \emph{near}-duplicates instead of \emph{exact} duplicates as page numbers are often
appended to running headers or footers by OCR\@; but we set the threshold conservatively to minimize
excess removal.
The top or bottom 5 lines of each page form the initial candidates for removal. We choose
the top and bottom 5 lines because OCR engines are not always consistent with placement of
marginalia and edge-matter; further, some books have multiline headers and footers.

For each of the top and bottom 5 lines of each page, we strip whitespace and discard candidates
shorter than 6 characters. Lines this short are not likely to be a running header/footer and
are too short to match reliably.
Each remaining candidate is represented as a MinHash~\parencite{broder2000identifying} signature made
from 128 permutations over character 5-grams.
Each signature is indexed in a MinHash locality-sensitive hashing (LSH) index\footnote{The headers and
footers are kept in separate LSH indices so that headers only match with other headers and footers
only match with other footers.} with Jaccard threshold $0.85$ (see for example Chapter 3
of~\cite{leskovec2020mining}; we use the implementation in the \textsc{datasketch} library
by~\cite{datasketch}).
To be marked as a duplicate, a line must have at least $3$ near-duplicates within a 5-page
span.\footnote{Headers and footers often differ between odd and even pages. To lower random matching, we require
not one repetition, but at least two. This explains the requirement for $\geq$3 instances within a
5-page span.}
These marked duplicates (in the top or bottom 5 lines, near-exact threshold, with at least 3 duplicates
in a cluster) are deleted from the middlematter.

\subsubsection{Results}

Running header and footer removal required 927h (combined wall-time across many 4-core jobs on Intel
Cascade Lake workstations) at an average of $3.4$s/volume. Removal statistics are given in
Table~\ref{table:header_footer}.
\begin{table}[htb!]
  {
  \centering
  \small
    \begin{tabular}{lc}
      \toprule
      Total Header and Footer Lines Removed & \numprint{272583638} \\
      Volumes with $\geq$1 Removal & \numprint{760238} (77.3\%) \\
      Average Removed/Volume & 277 \\
      \bottomrule
    \end{tabular}

  \vspace{1.5em}

  \begin{minipage}{0.45\linewidth}
    \centering
    \begin{tabular}{llc}
      \multicolumn{3}{c}{\footnotesize High Removal Languages} \\
      \midrule
      \textbf{Language} & \textbf{Avg/Book} & \textbf{Avg/Page} \\
      \midrule
      Scots & 233 & 0.93 \\
      English & 347 & 0.91 \\
      French & 289 & 0.66 \\
      Greek & 272 & 0.65 \\
      Latin & 221 & 0.59
    \end{tabular}
  \end{minipage}
  \hfill
  \begin{minipage}{0.45\linewidth}
    \centering
    \begin{tabular}{llc}
      \multicolumn{3}{c}{\footnotesize Low Removal Languages} \\
      \midrule
      \textbf{Language} & \textbf{Avg/Book} & \textbf{Avg/Page} \\
      \midrule
      Mandarin Chinese & 7.9 & 0.01 \\
      Slovak & 42.4 & 0.12 \\
      Ukrainian & 38.8 & 0.15 \\
      Arabic & 86.1 & 0.26 \\
      Gaelic & 48.3 & 0.27
    \end{tabular}
  \end{minipage}
  }
  \caption{Header and Footer Removal Statistics}\label{table:header_footer}
\end{table}
Random sampling shows that languages with vertical reading orientation have less accurate
header/footer detection. This appears to be primarily due to placement within the OCR text:
while we observed that headers and footers usually appear at the top or bottom of OCRed pages,
language-specific failure modes of OCR engines may lead to different, hard to anticipate placements.

\subsection{Page Number Removal}\label{ssec:pagenums}

As with running headers and footers, page numbers appear in margins and result in
artifacts at almost every page break. OCR engines typically place page numbers near the top or
bottom of each page. We look for short, numeric lines at the top and bottom of pages and remove
them.

\subsubsection{Methodology}

We detect a line as a page number if
\begin{enumerate}[nosep,noitemsep]
  \item it is in the top or bottom 5 lines of OCR output,
  \item it has at most 8 characters,
  \item at least one character is \emph{numeric}, and
  \item at most one non-whitespace character is \emph{non-numeric}.
\end{enumerate}
We consider a character \emph{numeric} if, after NFKC Unicode normalization, the character is in the
numeric Unicode category (specifically if
\mintinline{python}{unicodedata.category(ch).startswith('N')} is true in Python).

In addition, there is one later round of \emph{stray number removal} that occurs after separating
the text into sentences. Any ``sentence'' that consists of a single short number (with at most one
non-numeric character after removing white-space) is removed.

\subsubsection{Results}

Page number removal required a total of 2.5h (combined wall-time across many 4-core jobs on Intel
Cascade Lake workstations) at an average of less than 0.01 s/volume.
This is expected: page number detection is a short sequence of string inspections with no additional
models or hashing. Statistics for the run are given in Table~\ref{table:page_number}.
\begin{table}[htb!]
  {
  \centering
  \small
    \begin{tabular}{lc}
      \toprule
      Total Page Numbers Removed & \numprint{433722448} \\
      Volumes with $\geq$1 Removal & \numprint{980441} (99.7\%) \\
      Average Removed/Volume & 441 \\
      Average Removed/Page & 1.123 \\
      Median Removed/Page & 1.064 \\
      \bottomrule
    \end{tabular}
  \caption{Page Number Removal Statistics}\label{table:page_number}
  }
\end{table}
We observe that the median removal is close to one number per page, agreeing with expectation.
A common source of additional removals comes from footnote and caption markers, which are often
placed on their own lines by OCR engines.

\begin{figure}[htb!]
  \centering
  \includegraphics[width=0.8\textwidth]{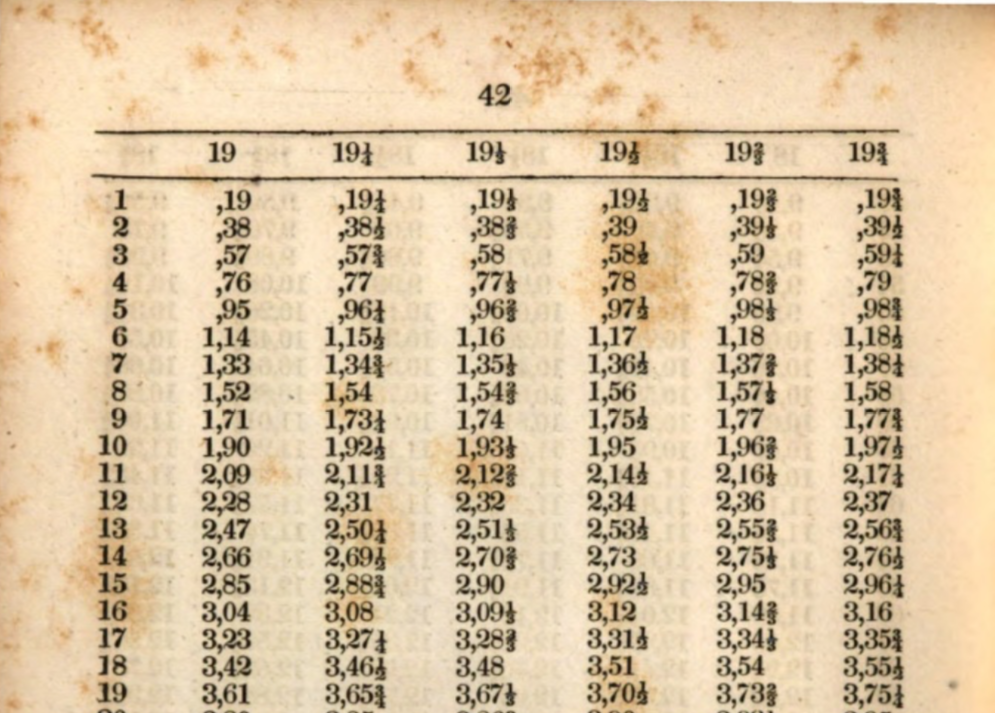}
  \captionsetup{width=0.8\linewidth}
  \caption{Part of a page from \textsc{HN3J4P}, \emph{The American ready reckoner}, showing how to
    multiply $19$, $19 {}^1{\mskip -2mu/\mskip -2mu}_4$, and so on by numbers from $1$ to $19$.
    Every table entry receives its own line in the OCRed output, so almost every checked line
    resembles a page number.}\label{figure:american_reckoner}
\end{figure}

The average is larger than the median. This reflects a long tail: 13k books had $\geq3$
page numbers removed per page.
The highest ratio is $8.7$ removed ``page numbers'' per page, held by the book with barcode
\textsc{HN3J4P}, titled \emph{The American ready reckoner: designed to insure correctness as well
as despatch in business}.
This book has a multiplication table on almost every page.
See Figure~\ref{figure:american_reckoner} for an example portion.
Currently available OCR text in IB-HL sometimes places every cell of a table on its own line.
Inspection shows that all top 10 books are similarly full of tables.

\subsection{Sentence Segmentation}\label{ssec:segment}

We now reorganize the middlematter into sentences.
The corpus spans roughly 250 languages with different punctuation conventions and scripts, so we use
multiple sentence segmentation strategies.

For languages with English-like punctuation, we use \textsc{Nupunkt}~\parencite{nupunkt_paper}, a modern
implementation of the Punkt algorithm~\parencite{punkt2006}. \textsc{Nupunkt} is an unsupervised algorithm that
attempts to recognize sentence-ending punctuation and common ways for that same
punctuation to occur without ending a sentence (such as in abbreviations
like \texttt{Dr.}, \texttt{U.S.}, etc.).

For other languages, we instead use \textsc{wtpsplit} and \textsc{SaT}
(Segment-any-Text, by~\cite{satmodels2024}), a multilingual sentence segmentation
system. Specifically, we use the
\textsc{sat-3l-sm}\footnote{\url{https://huggingface.co/segment-any-text/sat-3l-sm}} three-layer
general sentence segmentation model for its balance of speed and performance.\footnote{As
recommended by \textsc{SaT}, see
\url{https://github.com/segment-any-text/wtpsplit\#available-models}.}

Determining when to use \textsc{Nupunkt} and when to use \textsc{SaT} is not straightforward and is
described in more detail in Appendix~\ref{appendix:ss:segmentation}. The split is heavily skewed:
97.9\% of books go through the \textsc{Nupunkt} engine and the remaining 2.1\% go through the
\textsc{SaT} engine.

\subsubsection{Methodology}

For the 138 \textsc{Nupunkt}-compatible languages listed in Table~\ref{table:nupunkt_languages} in
the Appendix, the sentence segmentation process starts with a base corpus generated during
preprocessing (cf.\ \S\ref{ssec:preprocessing}) and adapts it for each book. In Punkt-based
algorithms, adaptation involves combining abbreviation lists and updating punctuation neighborhood
counts.

Combined base model training for \textsc{Nupunkt} required 3 hours (on 4 cores of a shared Intel
Cascade Lake cluster node).
It is possible to apply the base language \textsc{Nupunkt} models directly, without per-book
adaptation, but we found that performance improved with per-book adaptation. This was especially
true on books with many acronyms, abbreviations, and other technical non-sentence-terminating
punctuation. This comes with significant cost: per-book adaptation requires approximately $16\times$
more compute time than using only the base model to segment the book.\footnote{ For this reason, the
pipeline library allows simple configuration to enable or disable per-book \textsc{Nupunkt}
adaptation.} (Concretely: segmenting a book using a loaded model requires on average 0.6s on a
single core of an Intel Cascade Lake workstation; adapting a single book requires just over 10s on
average).

Punkt-based algorithms do not apply to every language, especially when punctuation
differs strongly from English. Determining when \textsc{Nupunkt} applies is complicated by the
continued incorporation of Western punctuation symbols into writing systems (see
e.g.~\cite{lee2014korean, twine1984adoption}).
Similar texts on similar subjects in the same language can have different punctuation standards.
For the 112 languages that we do not consider \textsc{Nupunkt}-compatible, we instead use the three-layer
general segmentation \textsc{SaT} neural network-based sentence segmenter.
The \textsc{SaT} model is the only step in the general pipeline that requires a
GPU\@. On an NVIDIA A100 GPU, segmenting a typical 400 page book takes 1.18s on average.

Thus \textsc{SaT} requires twice as much time as applying the base \textsc{Nupunkt} model, but is
\emph{faster} than applying per-book \textsc{Nupunkt} adaptation. On the other hand, \textsc{SaT} requires a
GPU\@.

\subsubsection{Results}

Sentence segmentation required a total of 2855h (combined wall-clock time on our distributed cluster;
heterogeneous nodes). See Table~\ref{table:nupunkt_timing} for timing by type. \textsc{Nupunkt}
adaptation and inference were performed on single-core Intel Cascade Lake workstation nodes.
\textsc{SaT} inference was performed on heterogeneous nodes containing at least 4 Intel Sapphire
Rapids CPU cores and 1 NVIDIA A100 GPU\@.
\begin{table}[htb!]
  \small
    \centering
    \begin{tabular}{llll}
      \textbf{Engine} & \textbf{Volumes} & \textbf{Total Time} & \textbf{Avg s/vol} \\
      \midrule
      \textsc{Nupunkt} Per-Book Adaptation & \numprint{962373} & \numprint{2691}h & 10.06s \\
      \textsc{Nupunkt} Inference & \numprint{962373} & 157h & 0.59s \\
      \textsc{SaT} (GPU) & \numprint{20629} & 6.6h & 1.18s
    \end{tabular}
    \captionsetup{width=0.8\linewidth}
    \caption{\textsc{Nupunkt} and \textsc{SaT} Timing Information.}\label{table:nupunkt_timing}
\end{table}

Output statistics are given in Table~\ref{table:nupunkt_stats}.\footnote{The number of words per sentence was
computed for this table using the
\textsc{polyglot} library.}
\textsc{Nupunkt} sentences and \textsc{SaT} sentences have approximately the same number of
characters, but \textsc{SaT} sentences tend to have far more words.
Whether this reflects language conventions or segmenter differences is unclear.

\begin{table}[htb!]
  \centering
  \small
    \begin{tabular}{lcccc}
      \textbf{Engine} & \textbf{Sentences} & \textbf{\% all sentences} & \textbf{Avg chars/sentence} & \textbf{Avg words/sentence} \\
      \midrule
      \textsc{Nupunkt} & \numprint{7010365214} & 99.00\% & 113.2 & 23.3 \\
      \textsc{SaT} & \numprint{70838120} & 1.00\% & 119.5 & 36.7
    \end{tabular}
  \caption{Sentence Segmentation Statistics}\label{table:nupunkt_stats}
\end{table}

\subsection{Topical Chunking}\label{ssec:chunk}

The next step of the pipeline is to organize the sequence of sentences into ``subtopic
paragraphs'' and ``subtopic sections''. Our motivation and terminology stem
from~\cite{hearst1997text}, which characterizes a text structure as a sequence of subtopics
within broader main-topic discussions.
We use ``subtopic paragraph'' and ``subtopic section'' to mean contiguous stretches of text that are
\emph{about} something. A single volume may contain many subtopic sections, and a single subtopic
section may contain many subtopic paragraphs.

This differs from the actual presentation within each volume. Organizing
sentences into hierarchical semantic sections is not how authors typically
organize their writing. To paraphrase Hearst, many volumes consist of long
sequences of paragraphs without any structural demarcation. Organizational
conventions change over time and vary across languages. We expect a more
consistent semantic structure to be valuable for downstream tasks. Where
sentences are too small and complete volumes are too big, subtopic paragraphs
and sections are self-contained and topically coherent. Subtopic chunks are
natural for context-aware search, semantic search and retrieval, long-context
focused LLM training, duplicate detection, and other information retrieval
tasks.

Our approach to topical chunking is a modern instantiation of Hearst's TextTiling algorithm. But
where TextTiling detects topical shifts from a variety of lexical patterns, we search for dips in
cosine similarity between static embeddings of sentences.

\subsubsection{Methodology}

We use the \textsc{Model2Vec} static model distilled from \textsc{BAAI/BGE-M3} (cf.\
\S\ref{ssec:preprocessing}) to generate static embeddings of sentences. The rest
of the approach is strongly informed by TextTiling~\parencite{hearst1997text}. Concretely:
\begin{enumerate}[nosep,noitemsep]
  \item Embed each sentence using a static model distilled from
    \textsc{BAAI/BGE-M3} (in batches of $256$ sentences).
  \item Score each sentence-gap by semantic cohesion. Take the mean embedding of
    the previous 5 sentences and the mean embedding of the next 5 sentences, and compute their
    cosine similarity. High scores suggest topical continuity. Low scores signal topical shift.
  \item Collect the sequence of gap scores into a similarity curve. We apply a single-pass moving
    average (width 3) to reduce high-frequency oscillation from single-sentence noise.
  \item Compute the TextTiling valley-depth at each gap. At each sentence-gap, we find the previous and
    subsequent peaks (where sentences are most topically cohesive) and sum the differences,
    \begin{equation*}
      \text{valley depth} = (\text{left\_peak} - \text{gap\_value}) + (\text{right\_peak} -
      \text{gap\_value}).
    \end{equation*}
    Deep, two-sided valleys correspond to strong topic boundaries. Shallow valleys do not.
  \item Greedily select sentence boundaries. A sentence-gap is a boundary candidate if its depth
    exceeds a per-book threshold $\bigl(\text{mean}(\text{valley\_depth}) -
    0.5\text{std}(\text{valley\_depth})\bigr)$. Candidates are taken greedily from deepest to
    shallowest, subject to the requirement that no segment be shorter than $3$
    sentences.\footnote{In the greedy selection algorithm, the last chunk in the book is not
    constrained to have at least $3$ sentences. The final chunk in each book may be smaller.}
  \item Use the accepted gaps as subtopic-paragraph boundaries.
\end{enumerate}

Our methodology deviates from Hearst in two places: we use static sentence embeddings
instead of term-frequency overlap and lexical statistics since sentence embeddings better
capture semantic content; and we operate on whole sentences instead of fixed-length pseudosentences
since sentence boundaries are already provided by the segmentation step in the pipeline.

Section-level chunking reuses \emph{exactly} this procedure, except using the identified
subtopic paragraphs in place of sentences. Iteration would yield larger tiers in the
semantic hierarchy.

\subsubsection{Results}

A total of \numprint{1568}h was required to perform subtopic chunking (combined wall-time across many
4-core jobs on Intel Cascade Lake workstations) with an average of $5.7$ s/volume. The median time
per book was $1.5$ s/volume and the overall time was heavily right-skewed by long books with many
sentences.
For a volume with $T$ tokens, $N$ sentences, and $B$ candidate paragraph breaks, our implementation
runs in time $O(T + N\log N + B^2)$; the terms come from embedding cost, sorting sentence gaps, and
greedily building the boundary list, respectively. In practice, $B \approx N/5$, hence long books
with many sentences will take a disproportionate amount of time. Core statistics are in
Table~\ref{table:chunk_stats}.
\begin{table}[htb!]
  \centering
  \small
    \begin{tabular}{lc}
      \toprule
      Total Sentences & \numprint{7081203334} \\
      Detected Subtopic Paragraphs & \numprint{1394049896} \\
      Detected Subtopic Sections & \numprint{297350019} \\
      Avg Paragraphs per Section & 4.688 \\
      Avg Sentences per Paragraph & 5.080 \\
      \bottomrule
    \end{tabular}
    \caption{Topical Chunking Statistics}\label{table:chunk_stats}
\end{table}
The average number of sentences per paragraph is consistent across languages: the smallest is
$\approx 4.8$ in Japanese, while the largest is $5.1$ in Latin. No significant difference in numbers
of sentences exists between volumes segmented using \textsc{Nupunkt} and volumes segmented using
\textsc{SaT}.

\begin{figure}[htb!]
  \begin{minipage}{0.28\textwidth}
    \centering \small
    \begin{tabular}{lc}
      \toprule
      Mean & 574.4 chars \\
      p25 & 211 chars \\
      Median & 435 chars \\
      p75 & 766 chars \\
      p99 & \numprint{2448} chars \\
      \bottomrule
    \end{tabular}
  \end{minipage}
  \hfill
  \begin{minipage}{0.68\textwidth}
    \centering
    \small
    \includegraphics[width=4.5in]{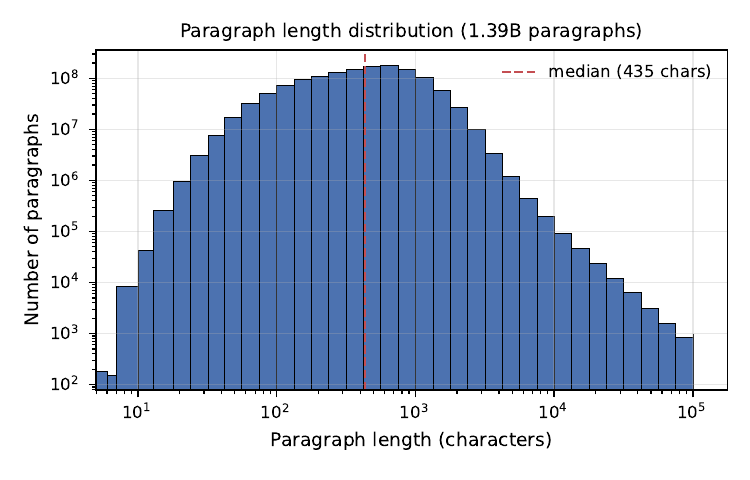}
  \end{minipage}
  \caption{Subtopic Paragraph Length Distribution}\label{figure:paragraph_length}
\end{figure}

The distribution of the lengths of paragraphs is shown in Figure~\ref{figure:paragraph_length}. The
plot shows a roughly log-normal, unimodal distribution. The 3-sentence minimum manifests as the
relatively sharp lower bound. There are 950 paragraphs with over 100k characters, mostly from poor
OCR on tables.

\subsection{Duplicate Identification}\label{ssec:deduplication}

Chunking the text into semantic subtopic paragraphs makes it easier to identify duplicate text.
We identify near-duplicate subtopic paragraphs across the whole dataset.
In general, they are not deleted\footnote{One exception comes from library boilerplate. Two clusters
consisting of library fine information are removed, as described in the Results section
\S\ref{sssec:duplication_results}.}
and are instead annotated and left in place.

Our treatment contrasts with common practice. Large-scale dataset preparation pipelines for language
model training routinely perform exact and fuzzy deduplication to remove repeated
documents~\parencite{brown2020language, gao2020pile}.
Some recent work suggests more rather than less aggressive
deduplication~\parencite{lee2022deduplicating}, and that even semantic near-duplicates can be
removed with little impact on downstream performance~\parencite{abbas2023semdedup}.

Though removing duplicate paragraphs might make sense for some tasks, it does not work for all
problems.
Instead, we mark paragraphs with duplicates so that downstream users can determine whether to use or
discard them as appropriate.

\subsubsection{Methodology}

Our approach to duplicate identification follows other large-scale deduplication efforts and reuses several
methodological components from page deduplication (cf.\ \S\ref{ssec:duplicate_page}).
We follow a typical simhash and LSH strategy over entropy-filtered 9-grams hashed with
MurmurHash3 and consider two simhashes duplicates if they differ in at most 5 bits.

We compute 128-bit simhashes~\parencite{charikar2002simhash} over 9-grams from hard-normalized (cf.\
\S\ref{ssec:unicode}), lower-cased text. Simhashes are only computed over 9-grams that contain at
least 4 distinct characters to avoid noise from strings of repeated text.
Each $9$-gram is hashed with MurmurHash3 and accumulated into a standard
simhash.\footnote{We again use a small C++ extension for computing simhashes. See
Appendix~\ref{appendix:ss:duplicate_page}.}

Thus each of the $\approx$1.4B paragraphs is reduced to a 128-bit signature. Two simhashes that
differ by at most 5 bits are considered duplicates. This is conservative but allows small
typographic deviations or OCR artifacts to not disrupt duplicate identification. The naive strategy of
pairwise comparison (as executed with page deduplication) is no longer viable.
Instead, to identify these duplicate simhashes, each 128-bit signature is split into 6 bands (with
$(22, 21, 21, 21, 21, 22)$ bits, respectively). If two signatures differ by at most 5 bits, then
they must exactly agree on at least one band. Thus we seek signatures that exactly match at least
one band.

We simultaneously determine all matches in a band by sorting a document with lines of the form
\texttt{(band\_value) | doc\_id}. Each contiguous range of identical band values is one bucket of
candidates to check for duplicates. We create 6 documents (one for each band) and sort each
individually. The natural unique identifier for each document is its
\texttt{barcode.paragraph\_index}. As many barcodes are 14-character strings, if we used this
identifier then the band-file to sort would have size
\begin{equation*}
  \approx 1.4 \times 10^9 \times (3 \; \text{bytes} + 18 \; \text{bytes}) \approx 30\text{GB}.
\end{equation*}
This would be larger than necessary and negatively affect sort time.
Instead, we construct a new document id for efficient (and predictable) byte packing.

After creating simhashes for all paragraphs, we read these simhashes in a fixed order and assemble

\begin{enumerate}[nosep,noitemsep]
  \item A global binary simhash array \texttt{hashes.bin}, the global concatenation of all paragraph
    simhashes. (1.4B paragraphs, each with a 16-byte simhash, yielding a file of size $\approx$23GB)
  \item An auxiliary array \texttt{book\_ids}, a list of volume barcodes in simhash-reading-order.
    (1M strings, tens of MB in size)
  \item An auxiliary array \texttt{book\_offsets}, one \texttt{uint64} per volume giving the index in
    \texttt{hashes.bin} at which that volume's paragraphs begin. (1M \texttt{uint64} integers, 8MB in
    size).
\end{enumerate}
These are constructed in order, so \texttt{book\_offsets} strictly increases and allows one to
detect both book starts and book ends; and the $n$th entry in \texttt{book\_offsets}
corresponds to the $n$th barcode in \texttt{book\_ids}.
With these arrays, the global document index of a paragraph is \emph{its index in the global simhash
array}. Binary search on \texttt{book\_offsets} allows recovery of the barcode and paragraph index.
We add a byte-packing assumption that the document index can be specified by one
\texttt{uint32} for minor convenience (capping tractable corpus size to $2^{32} \approx 4.29$B
paragraphs).

Thus each paragraph is uniquely specified by a 4-byte \texttt{doc\_id}. We set up band sorting by
packing each \texttt{(band value, doc\_id)} into a single \texttt{uint64} key
{\setlength{\topsep}{0pt}\setlength{\partopsep}{0pt}
\begin{verbatim}
                    key = (band_value << 32) | doc_id.
\end{verbatim}
}
The band-files for sorting are thus each about 11GB in size and consist of 64-bit integers for
sorting. These can be sorted in memory and a single linear scan recovers all candidate buckets.

To process the buckets in parallel, we memory-map the global simhash array \texttt{hashes.bin}. A
pool of workers computes exact pairwise Hamming distances using two
\texttt{xor}s, two \texttt{popcnt}s, and an addition to combine. Positively identified results are
added to a global union-find data structure, which ultimately yields the clusters.

After clusters are identified, a unique \texttt{cluster\_id} is assigned (of the form
\texttt{barcode:par\_idx}, chosen as the alphabetically first barcode of a volume in the cluster).
During annotation, every paragraph in a cluster is annotated with one of two tags: the paragraph
that was chosen as the \texttt{cluster\_id} receives the annotation
{\setlength{\topsep}{0pt}\setlength{\partopsep}{0pt}
\begin{verbatim}
    <p data-clusterid="BARCODE:PAR_IDX" data-representative>
    ...
    </p>
\end{verbatim}
}
and non-representative paragraphs are wrapped in an aside tag
{\setlength{\topsep}{0pt}\setlength{\partopsep}{0pt}
\begin{verbatim}
    <aside data-cluster="BARCODE:PAR_IDX">
    <p> ... </p>
    </aside>
\end{verbatim}
}
We describe this more when discussing annotation in \S\ref{ssec:annotation}.

\subsubsection{Results}\label{sssec:duplication_results}

Duplicate paragraph identification required 129h (127h of combined wall-clock time across distributed
nodes with 4 Intel Cascade Lake cores to compute simhashes, and 1.5h on a 16-core MacBook Pro M4 Max
for simhash comparison).
The statistics in Table~\ref{table:duplicate_stats} show that
\numprint{1932557} same-book duplicate clusters were detected, \emph{fewer} than the number
of removed duplicate pages (\numprint{1981296} as discussed in
\S\ref{sssec:duplicate_page_results}; removing duplicate pages thus significantly reduced
unnecessary duplicate identification).
Though some volumes repeat paragraphs internally, most detected duplicates span multiple volumes.

\begin{table}[htb!]
  \centering
  \small
    \begin{minipage}{0.6\textwidth}
      \centering
    \begin{tabular}{lc}
      \multicolumn{2}{c}{\textbf{Identified Duplicate Counts}}\\
      \midrule
      Total Subtopic Paragraphs & \numprint{1394049896} \\
      Duplicate Representatives & \numprint{51233629} \\
      Duplicate Non-Representatives & \numprint{72856480} \\
      Average Cluster Size & 2.42 paragraphs \\
      Single-book Clusters & \numprint{1932557} (3.8\%) \\
      Multi-book Clusters & \numprint{49301072} (96.2\%)\\
      Books With Identified Duplicates & \numprint{557961} (56.8\%) \\
    \end{tabular}
    \end{minipage}
    \begin{minipage}{0.35\textwidth}
      \centering
      \begin{tabular}{cc}
        \textbf{Cluster Size} & \textbf{\# Clusters} \\
        \midrule
        2 & \numprint{40123977} \\
        3 & \numprint{7077321} \\
        4 & \numprint{2119430} \\
        5--10 & \numprint{1725371} \\
        11--100 & \numprint{186069} \\
        101--1000 & \numprint{1376} \\
        1000+ & \numprint{85}
      \end{tabular}
    \end{minipage}

    \caption{Duplicate Identification Statistics}\label{table:duplicate_stats}
  \end{table}

  \begin{table}[htb!]
    \centering
    \small
    \begin{tabular}{ccc}
      \textbf{Language} & \textbf{\# Duplicate Paragraphs} & \textbf{\% of All Paragraphs in Language} \\
      \midrule
      English & \numprint{57815405}     & 8.09\%   \\
      German & \numprint{5976526}      & 2.41\%   \\
      French & \numprint{5804669}      & 3.07\%   \\
      Latin & \numprint{957619}        & 2.26\%  \\
      Italian & \numprint{640421}        & 1.38\%  \\
      Spanish & \numprint{475111}        & 1.80\%  \\
      Dutch & \numprint{219803}        & 1.32\%  \\
      Russian & \numprint{153856}        & 0.61\%  \\
      Hungarian & \numprint{47632}         & 0.75\%  \\
      Hebrew & \numprint{39949}         & 1.04\%  \\
    \end{tabular}

    \caption{Language Distribution of Detected Duplicate Paragraphs}\label{table:duplicate_stats2}
\end{table}
Table~\ref{table:duplicate_stats2} shows English as a pronounced outlier in
duplication rate, at more than $2.5$ times the rate of the next language.
This likely reflects the size of the English subcollection: with 487k English volumes, there is a
bigger pool of potential duplicate books (including reprints, reorganized collections, etc.).

The average cluster consists of $\approx$2.42 duplicates, but the distribution is long-tailed.
We found 85 clusters containing over 1000 different duplicates. Inspection shows that these
exceptionally large clusters tend to be either missed boilerplate or poorly-OCRed tabular filler.
The two largest were variations of the same library fine description: 
{\setlength{\topsep}{0pt}\setlength{\partopsep}{0pt}
{\small
\begin{verbatim}
    REP  32044004341889.672  - 21,056 Copies
    REP 32044004442943.3353  - 19,072 Copies
    ----------------------------------------
    This book should be returned to the Library on or before the
    last date stamped below. A fine of five cents a day is incurred
    by retaining it beyond the specified time. Please return promptly.
\end{verbatim}
} }
As these were missed boilerplate (and were always the exact last paragraph in the book, from the
back cover), we removed these paragraphs from the dataset.

The next several largest clusters are all tabular noise:

\begin{minipage}{\textwidth} 
  {\setlength{\topsep}{0pt}\setlength{\partopsep}{0pt}
  {\small
  \begin{verbatim}
  REP  32044000056234:311  | 16,954 |  Do. Do. Do. Do. Do. Do. Do. Do. Do. Do.
  REP  32044001172907:42   | 16,804 |  ……………… …………..……………….. ………………
  REP  32044004599593:258  | 14,942 |  Dollars. Dollars. Dollars. Dollars.
  REP  32044010095289:65   | 13,360 |  idem. idem. idem. idem. idem. idem. idem.
  REP  32044002052694:1091 | 12,708 |  do. do. do. do.
  REP  32044004778643:383  |  9,033 |  Value. Quantity. Value.
  REP  32044004365193:2145 |  7,896 |  Do. Do. Do.
  REP  32044004554093:4597 |  7,262 |  ..do. .do. .do. ..do.
  \end{verbatim}
  }
  }
\end{minipage}
The \texttt{Do. Do.} repetitions are shorthand for ``ditto'', the same as \texttt{idem.
idem.} The repeated \texttt{Dollars.} and \texttt{Value. Quantity.} lines are orphaned lines
from tables.

\begin{figure}[htb!]
  \centering
  \includegraphics[width=0.9\textwidth]{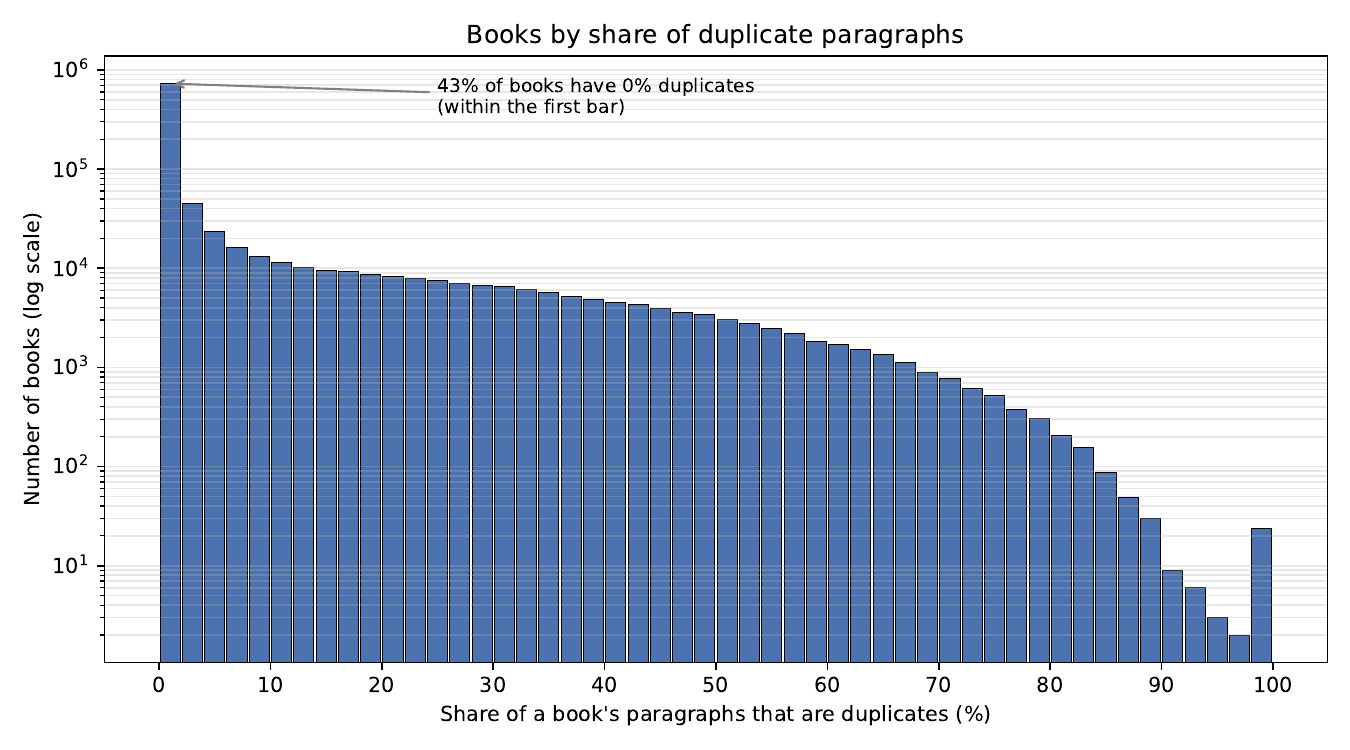}
  \caption{Books by \% Duplicate Subtopic Paragraphs}\label{figure:duplicate_fraction}
\end{figure}

Figure~\ref{figure:duplicate_fraction} shows the number of books binned by the
fraction of identified duplicate subtopic paragraphs. About 43\% of books have no duplicate
paragraphs and 80\% are under 5\%.
The figure shows a smooth $S$-curve of decay, except for
24 books with 98--100\% duplication. Inspection shows that these 24 books are
reprintings, edition changes, or duplicate books in the collection. For example, the three volumes
with barcodes \textsc{HWJX6S}, \textsc{HWJX7H}, \textsc{HWJX63} all contain a play \emph{The Pigeon}
by John Galsworthy, which is included in the 3-play anthology \textsc{32044086883501}.

\subsection{Bits-Per-Byte}\label{ssec:bpb}

OCR quality over a heterogeneous, multi-century, multilingual collection produces text of varying
data fidelity. Earlier stages of the pipeline have shown that OCR quality is generally high. But we
have also observed that tables and marginalia are often poorly OCRed and yield noisy paragraphs in
this collection.

We seek a computable, language-agnostic number per paragraph that:
\begin{enumerate}[nosep,noitemsep]
  \item ranks paragraphs by how predictable, fluent, and well-formed their text is,
  \item can be summarized per book as a quality statistic, and
  \item facilitates downstream filters that keep typical prose and omit
    unusual extremes.
\end{enumerate}
One approach is to use language model \emph{cross-entropy}. It is common to normalize
by the number of tokens and report the exponentiated value as \emph{perplexity}, which has a
long history as a corpus-filtering signal (e.g.~\cite{wenzek2020ccnet},~\cite{gao2020pile},
and~\cite{marion2023less}). But the numeric scale of perplexity depends both on the underlying model
tokenizer and the language of the text; tokenizers can produce different numbers of tokens for
different scripts as a result of different priorities and availability of training data. We instead
normalize by bytes: the Bits-Per-Byte (BPB) measures how many bits the model needs to encode each
byte of text, which is closer to tokenizer-independent, language-independent data.

Computing BPB is more expensive than the rest of the pipeline and requires
dedicated GPU resources. We think of computing BPB as an optional
phase performed after the primary processing pipeline. Our implementation allows
simple configuration to skip BPB evaluation. But as per-paragraph BPB enables
users to filter or seek out-of-distribution text without needing to perform
these computations themselves, we believe this is a valuable contribution.

\subsubsection{Methodology}

For each subtopic paragraph, we compute its BPB with a causal language model and accumulate the
next-token log-likelihood it assigns, then divide by byte count:
\begin{equation*}
  \text{BPB} = \frac{1}{\ln 2} \cdot \frac{-\sum_t \ln p_{\theta}(x_t \mid x_{<t})}{\text{(number of UTF-8 bytes)}}.
\end{equation*}
We use \textsc{Qwen/Qwen3-0.6B-Base}~\parencite{qwen3technicalreport} as the reference model
as it has strong multilingual support. We use a relatively small 0.6B parameter model so that
it is computationally possible to compute BPB for all 1.4B paragraphs in the
collection.\footnote{In Appendix~\ref{appendix:ss:bpb} we show that
  \textsc{Qwen/Qwen3-0.6B-Base} is an effective proxy for approximating
perplexity and BPB for larger \textsc{Qwen} models.}
The numerator is exactly the model's total negative log-likelihood of the paragraph. Lower BPB means
the paragraph is more predictable; higher BPB means it is more surprising.

Prior to computing BPB, we sort all paragraphs in each book by length and create batches of 16
paragraphs that all have approximately the same length.
As shown in Figure~\ref{figure:paragraph_length}, the distribution of paragraph lengths is
roughly log-normal with a long tail; thus presorting prevents large amounts of
unnecessary padding in the computation.

In our implementation, excessively short and excessively long paragraphs are each assigned a
sentinel value of $-1$ rather than a score. Paragraphs with $\leq 4$ characters are too short for
meaningful contextual prediction. And paragraphs longer than $\numprint{14134}$ tokens also receive
the $-1$ sentinel value. This is due to our implementation encountering a 32-bit index limit in the
PyTorch CUDA kernel. The number of logits produced from a paragraph is \texttt{num\_tokens x
vocab\_size} and must have an \texttt{int32} address. The \texttt{vocab\_size} in our model is
\numprint{151936}, leaving $(2^{31} - 1)/151936 \approx 14134$ as the max size. In addition, if the
BPB computation leads to a GPU Out-Of-Memory error, the paragraph is assigned a $-1$ sentinel value.
This latter case occurred on 187k paragraphs during the initial pipeline run; these were resolved
by recomputation and the released dataset contains no such cases.

We compute some summary statistics about the BPB distribution across subtopic paragraphs for each
book: the min, max, mean, median, and the 10th, 30th, 70th, and 90th percentiles. Each of these
statistics ignores paragraphs with the sentinel value $-1$. Further, in the final dataset we do not
include BPB annotations with the sentinel value.

We compute these statistics at the book level because absolute BPB values depend
strongly on genre and technical material.
BPB should be interpreted as a measure of predictability and not of
value. Legitimate-but-unusual content, including mathematical notation or
low-resource languages underrepresented in the language model's training, will
be ``surprising'' to the model and have a high BPB value. By providing per-book
statistical distributions, we encourage relative filtering instead of
establishing corpus-wide cutoffs.

\subsubsection{Results}

Computing BPB for all 1.4B paragraphs in the collection required a total of $\approx2000$ GPU-hours
($\approx$83 GPU days, combined wall time over hundreds of GPU nodes on a distributed, heterogeneous
cluster; each node had one NVIDIA A100 GPU and at least 8 Intel Sapphire Rapids CPU cores) with an
average of $\approx7$ s/volume or $\approx$700k paragraphs per GPU-hour.

\begin{table}[htb!]
  \centering
  {\small
    \begin{tabular}{lccccc}
      \textbf{Subcollection} & \textbf{\# Subtopic Paragraphs} & \textbf{Mean} & \textbf{p25} & \textbf{Median} &\textbf{p75} \\
      \midrule
      All$^{\dagger}$                    &  \numprint{1393997373}  &  1.666  &  1.179  &  1.510   &  2.033 \\
      \textsc{Nupunkt}        &  \numprint{1379804105}  &  1.661  &  1.178  &  1.506   &  2.026 \\
      \textsc{SaT}            &  \numprint{14193268}    &  2.179  &  1.532  &  2.076   &  2.824
    \end{tabular}
  }

  \vspace*{0.7em}

  \parbox{0.85\textwidth}{\footnotesize $^\dagger$The \numprint{1394049896} subtopic paragraphs in
  Table~\ref{table:chunk_stats} are reduced to \numprint{1394009768} in the released dataset after
  removing \numprint{40128} paragraphs in the two deleted library-fine boilerplate clusters
  (\S\ref{sssec:duplication_results}). Of these, \numprint{1393997373} receive a BPB score and
  \numprint{12395} keep the $-1$ sentinel. This leads to the small differences in totals in these
tables.}
  \caption{Core BPB Statistics}\label{table:basic_bpb_stats}
\end{table}

\begin{figure}[htb!]
  \centering
  \fbox{\includegraphics[width=0.9\textwidth]{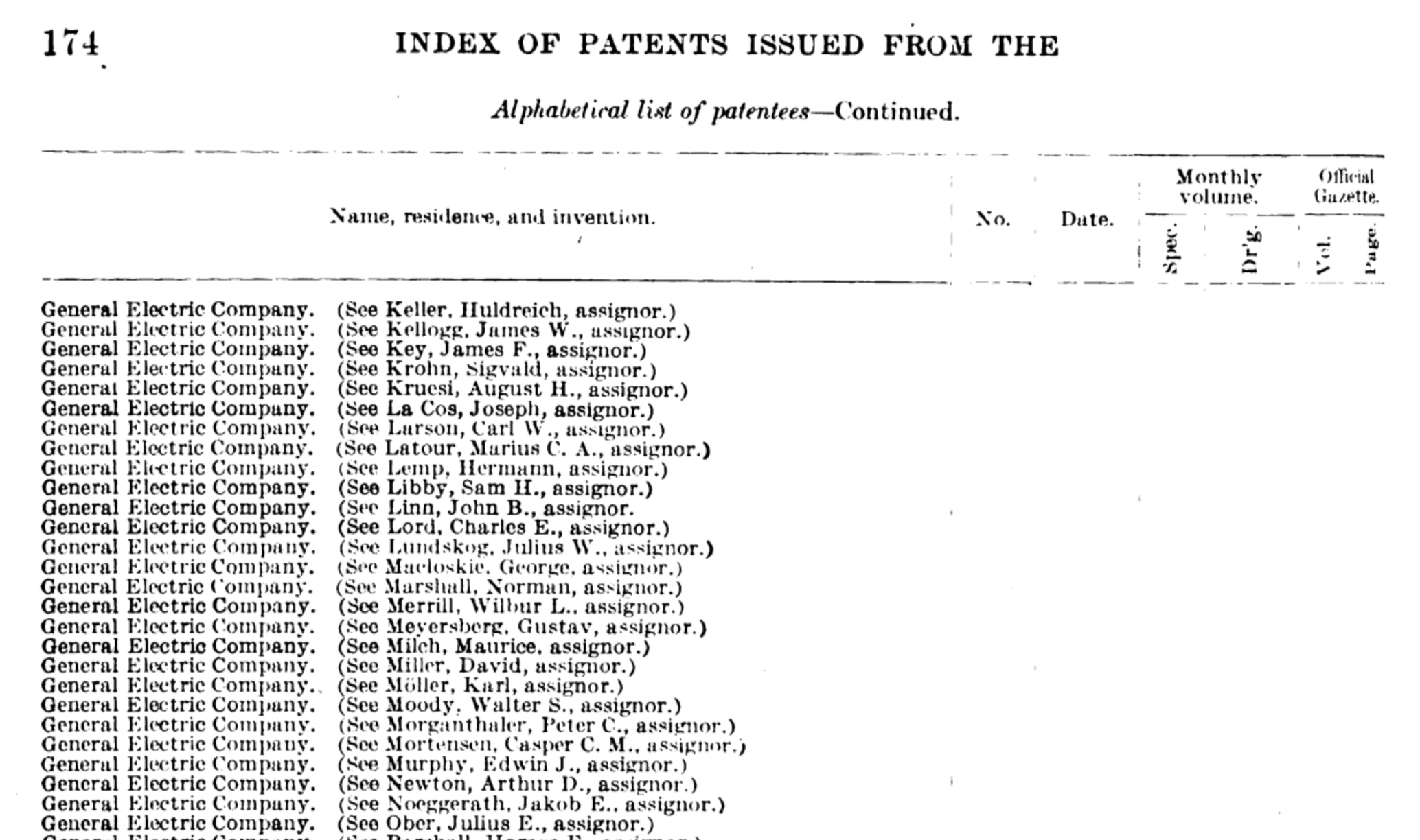}}
  \captionsetup{width=0.8\linewidth}
  \caption{Book \textsc{HJ1A7X} (\emph{Annual Report on Patents for the US House of Representatives,
  1906}) containing the ``paragraph'' with the smallest bits-per-byte.}\label{figure:patent_book}
\end{figure}

See Table~\ref{table:basic_bpb_stats} for statistics on the distribution.
\textsc{Nupunkt}-compatible languages have smaller BPB on average. One possible explanation for this
difference is that \textsc{Nupunkt}-compatible languages (such as English) form the bulk of the
pre-training data for the \textsc{Qwen3} language model. A second possibility is that
currently available OCR is more accurate on \textsc{Nupunkt}-compatible languages than on
languages with different scripts and punctuation.

The single paragraph with the smallest (valid) BPB is from a book of patent records and is shown in
Figure~\ref{figure:patent_book}. The OCR engine parsed the column of ``General Electric Company.''
as a single block of text and the identified subtopic paragraph consists of several repetitions of
the text ``General Electric Company.'' According to the \textsc{Qwen3-0.6B-Base} model, this
paragraph has BPB value 0.0297, reflecting its repetition.

The paragraphs identified with the largest BPB appear to consist mostly of OCR garbled text.
These extreme examples confirm that BPB can be a helpful predictor of text
quality and might be useful when filtering the data.

A total of \numprint{12395} paragraphs ($\approx$0.0009\%) carry the $-1$ sentinel in the released
dataset. Of these, \numprint{12391} are longer than the \numprint{14134} token limit and $4$ are too
short (which are all OCR noise on the last page of books).

\subsection{Final Annotation and Formatting}\label{ssec:annotation}

The final step of the pipeline is to collect the processed middlematter, the separated endmatter,
the duplicate identification clusters, and the per-subtopic paragraph BPB values together to
generate the enriched text output. In addition, we compute and incorporate some metadata.

\paragraph{Output Format.}
The pipeline emits three annotated strings per volume: \texttt{frontmatter\_gen},
\texttt{middlematter\_gen}, and \texttt{backmatter\_gen}. Each is assembled from a small set of
HTML-like tags and consists of HTML-escaped text. We choose HTML due to the availability of parsing
tools and standards (though we also release a small, custom parser library, cf.\
Appendix~\ref{appendix:ibet_parser}). HTML allows separation of annotation and primary content:
this is the difference between HTML attributes/structural elements and inner text.

Inside the endmatter, each non-empty page becomes one \texttt{<div>}. There is exactly one
\texttt{<div>} per endmatter page and empty pages are discarded. Each \texttt{<div>} is annotated by
its detected class, taking the form
{\setlength{\topsep}{0pt}\setlength{\partopsep}{0pt}
\begin{verbatim}
      <div class="toc_index"> ... page text ... </div>
      <div class="biblio"> ... page text ... </div>
      <div class="otherendmatter"> ... page text ... </div>
\end{verbatim}
}

The middlematter is grouped into subtopic sections, each of which consists of subtopic paragraphs.
Each subtopic section is demarcated with a \texttt{<section>} tag. When available, each section tag
carries a \texttt{data-bpb} annotation with the mean BPB of the section (computed as the
non-weighted average BPB of the paragraphs in that section). This takes the form
{\setlength{\topsep}{0pt}\setlength{\partopsep}{0pt}
\begin{verbatim}
      <section data-bpb="X.XXXX"> ... paragraphs ... </section>
\end{verbatim}
}
The BPB values are given to 4 decimal digits.

Each subtopic paragraph is indicated with a \texttt{<p>} tag and can carry up to four attributes:
\begin{itemize}[nosep,noitemsep]
  \item \texttt{data-bpb="X.XXXX"}: per-paragraph bits-per-byte to 4 decimal digits. This is omitted
    when the computed BPB is $\leq0$ (i.e.\ when it has the $-1$ sentinel value).
  \item \texttt{data-language="LAN"}: per-paragraph detected language in ISO-639-3 from
    the \textsc{polyglot} library. Can be \texttt{UNKNOWN} and is omitted if \texttt{None}.
  \item \texttt{data-representative}: a Boolean flag indicating that this paragraph is the chosen
    cluster id paragraph among a cluster of identified duplicate paragraphs (cf.\
    \S\ref{ssec:deduplication}). If present, this indicates that this paragraph has the volume with
    the alphabetically-first barcode among a class of duplicate paragraphs.
  \item \texttt{data-clusterid="BARCODE:PARIDX"}: when \texttt{data-representative} is present, this
    attribute will be present and gives the name of the cluster. 
\end{itemize}
In addition, if a paragraph is identified as a duplicate of a \emph{different} paragraph, then the
paragraph is wrapped in an \texttt{aside} tag with a \texttt{data-cluster="BARCODE:PARINDICES"}
attribute. If it is a single paragraph, the \texttt{data-cluster} takes the form
\texttt{BARCODE:PARIDX}. But often a contiguous sequence of paragraphs duplicates another contiguous
sequence of paragraphs. When this occurs, the \texttt{data-cluster} takes the form
\texttt{BARCODE:N-M} where \texttt{N} and \texttt{M} are the first and last indices of the cluster
source.
Prototypical output resembles the following (with whitespace here emphasizing annotation structure):
{\setlength{\topsep}{0pt}\setlength{\partopsep}{0pt}
\begin{verbatim}
      <section data-bpb="X.XXXX">
        <p data-bpb="X.XXXX" data-language="LANG"
           data-representative data-clusterid="ABCDEF:15">
           ... source text for a duplicate class ...
        </p>
        <p data-bpb="X.XXXX" data-language="LANG">
          ... paragraph text ...
        </p>
        <aside data-cluster="GHIJKL:21">
          <p data-bpb="X.XXXX" data-language="LANG">
           ... duplicate text ...
          </p>
        </aside>
      </section>
      <section>
        ...
      </section>
\end{verbatim}
}

Stripping all HTML tags and parsing only HTML inner text recovers the volume text.
However, the HTML attributes encode metadata that might be useful for filtering or structured reasoning.

The following 21 metadata descriptors are included:
\begin{center}
  \small
  \begin{tabular}{lll}
    \texttt{bpb\_min\_gen}                 &   \texttt{bpb\_p10\_gen}                 &   \texttt{bpb\_p30\_gen} \\
    \texttt{bpb\_median\_gen}              &   \texttt{bpb\_p70\_gen}                 &   \texttt{bpb\_p90\_gen} \\
    \texttt{bpb\_max\_gen}                 &   \texttt{bpb\_avg\_gen}                 &   \texttt{primary\_language\_gen}  \\
    \texttt{language\_distribution\_gen}   &   \texttt{token\_count\_gen}             &   \texttt{char\_count\_gen} \\
    \texttt{word\_count\_gen}              &   \texttt{sentence\_count\_gen}          &   \texttt{paragraph\_count\_gen}  \\
    \texttt{section\_count\_gen}           &   \texttt{bigram\_count\_gen}            &   \texttt{bigram\_count\_unique\_gen}  \\
    \texttt{trigram\_count\_gen}           &   \texttt{trigram\_count\_unique\_gen}   &   \texttt{tokenizability\_ratio\_gen}
  \end{tabular}
\end{center}
We use the suffix ``\texttt{\_gen}'' to indicate values computed by this
pipeline, including all these metadata descriptors.
The \texttt{bpb\_XXX\_gen} metadata contain the min, max, average, median, 10th percentile, 30th
percentile, 70th percentile, and 90th percentile among the bits-per-byte values of subtopic
paragraphs in that volume.
The languages and proportions are encoded as
\begin{itemize}[nosep,noitemsep]
  \item \texttt{primary\_language\_gen}: the primary language code (given as ISO 639-3
    names)\footnote{\url{https://www.iso.org/iso-639-language-code}} as reported in IB-HL.
  \item \texttt{language\_distribution\_gen}: a dictionary of \texttt{(lang: proportion)} pairs,
    where \texttt{lang} is an ISO 639-3 language code and \texttt{proportion} is
    a float from $0.0$ to $1.0$ describing the proportion of the paragraphs in
    that language in the book.
\end{itemize}
The remaining metadata follows the same naming convention as in IB-HL (see \S4.8 of
\cite{cargnelutti2025institutional}).

\begin{sloppypar}
Finally, we include one filtered plaintext version of the collection called
\texttt{processed\_middlematter\_gen}. This is a plaintext version that skips duplicate paragraphs
(keeping only the representative cluster id for each class), keeps only paragraphs with paragraph-level
BPB between the 10th and 90th volume-level percentile, and otherwise removes all other annotations.
We provide this as an easy way to directly interact with a filtered version of
the dataset. This serves as an opinionated baseline for users to compare against
when parsing and filtering the dataset.
\end{sloppypar}

\subsubsection{Methodology}

All volume text is HTML-escaped using \texttt{html.escape(quote=False)} in Python during assembly.
The remaining work consists of computing additional metadata and standard string formatting.

In IB-HL, effort was made to identify languages on chunks of text up to 768 characters long (cf.\
\S4.4.1 of~\cite{cargnelutti2025institutional}). This provided information on the language
distribution within each volume. In this pipeline, we run a language-detection algorithm on every
subtopic paragraph. As subtopic paragraphs correspond more closely to semantic chunks, this should
give a more granular view of language distribution.

We use the \textsc{polyglot}\footnote{\url{https://github.com/aboSamoor/polyglot}}
library~\parencite{polyglot} to detect the language of each subtopic paragraph with one minor
modification: if \textsc{polyglot} fails to detect the language for a particular paragraph, and that
paragraph has fewer than $30$ \textsc{o200k\_base} tokens (computed using~\parencite{tiktoken}),
\emph{and} the two surrounding paragraphs had positive language detections with the same language,
then we assume the middle paragraph also has that same language. Language detection accuracy
degrades on short texts; however, contiguous chunks are typically in the same language.

For the rest of the metadata: token count and tokenizability are computed using
\textsc{tiktoken}~\parencite{tiktoken} with respect to the \textsc{o200k\_base}
tokenizer. ``Tokenizability'' is a score that measures how efficiently
\textsc{o200k\_base} can encode the text; in particular, it measures how close
to 1.25 tokens per word the text is. Word counts and lists are computed using \textsc{polyglot}.
Sentence counts, paragraph counts, and section counts are computed using the segmentation and
chunking from above. Bigrams and trigrams are computed in pure Python over the \textsc{polyglot}
word lists.

\subsubsection{Results}

Metadata computation required approximately 150h (combined wall-time on a 16-core MacBook Pro M4
Max). Exact timing was not logged.
Table~\ref{table:parlangs} contains detected language counts. This table contains the top ten most
commonly detected languages in the collection and the number of subtopic paragraphs detected in that
language. For comparison, we also present the number of volumes having that language as the dominant
language according to IB-HL.
\begin{table}[htb!]
  \centering
  {\footnotesize
  \begin{tabular}{ccccc}
    \textbf{Language}    & {\bf\shortstack{Detected Subtopic Pars \\ in Language}}  & \textbf{(\% of All Pars)} &
    {\bf\shortstack{Volumes with this\\Primary Language}}& \textbf{(\% of All Volumes)}\\
    \midrule
    English &           \numprint{732171247} & 52.52\% &           \numprint{487353} & 49.58\%  \\
    German  &           \numprint{221319743} & 15.88\% &           \numprint{157776} & 16.05\%  \\
    French  &           \numprint{174288052} & 12.50\% &           \numprint{135872} & 13.82\%  \\
    Latin   &            \numprint{51553325} &  3.70\% &            \numprint{21640} &  2.20\%  \\
    Italian &            \numprint{42287689} &  3.03\% &            \numprint{46074} &  4.69\%  \\
    Spanish &            \numprint{23468303} &  1.68\% &            \numprint{28836} &  2.93\%  \\
    Russian &            \numprint{21807544} &  1.56\% &            \numprint{15115} &  1.54\%  \\
    Dutch   &            \numprint{15615935} &  1.12\% &            \numprint{12682} &  1.29\%  \\
    Greek   &            \numprint{13726684} &  0.98\% &             \numprint{6319} &  0.64\%  \\
    Danish  &            \numprint{13142232} &  0.94\% &             \numprint{7899} &  0.80\%
  \end{tabular}
  }
  \caption{Paragraph Language Data}\label{table:parlangs}
\end{table}

The per-paragraph distribution closely agrees with the per-book distribution for large languages.
We observe that Latin (\texttt{lat}) has only $2.20\%$ of books in IB-HL and $3.70\%$ of paragraphs
across the collection. Many non-Latin books include quotes or passages in Latin.

As partial validation, we compared the detected primary language from IB-HL against the most common
language (by paragraph count) detected in each volume in this pipeline. We found
98\% agreement, and volumes with differing identified languages tended to be either bilingual or in
pairs of languages with large overlaps in n-grams, such as German-Yiddish-Dutch or Occitan-Catalan
or Montenegrin-Croatian.

\section{Dataset Analysis}\label{sec:analysis}

We present three focused analyses on separate aspects of the pipeline. Each analysis serves to
validate part of the pipeline and to illustrate research directions facilitated by \ibet{}. First,
we compare text statistics of \ibet{} against IB-HL to quantify the effect of cleaning. Second, we
combine duplicate identification with language detection to measure reuse \emph{across} languages.
Third, we pair BPB with publication metadata to trace how text predictability changes across the
Harvard Library book collection. Each of these analyses depends on a different annotation layer
added in this pipeline.

\subsection{Tokenizability in IB-HL and \ibet{}}\label{ssec:token_comparison}

Constructing \ibet{} extends the text-refinement process begun in IB-HL. We quantify
the differences between this dataset and IB-HL by comparing statistics.
Table~\ref{table:ib1_v_ibet_1} contains character, bigram, trigram, and tokenizability statistics.
(The rows for \ibet{} correspond to the final middlematter with all annotations removed).
The \ibet{} dataset has $\approx$26B fewer characters than IB-HL; this is primarily due to
separating endmatter, though running header/footer removal and page number removal have a small
contribution. The bigram and trigram unique ratios are smaller by at least 5 percentage points,
while tokenizability increases by $\approx$6 points. These are all consistent with less noise and
effective cleaning.

\begin{table}[htb!]
  \centering
  {
  \small
  \setlength{\tabcolsep}{5pt}
  \begin{tabular}{@{}lcccc@{}}
    \textbf{Dataset} & \textbf{\# Characters} & \textbf{Bigram uniq \%} & \textbf{Trigram uniq \%} & \textbf{Tokenizability} \\
    \midrule
    IB-HL (src OCR) & 828.48B  &  46.40\%  & 74.46\%  & 80.43 \\
    \ibet{} & 802.11B & 38.79\% & 69.23\% & 86.57 \\
    \midrule
    IB-HL (Top-5 Langs)  & 729.65B & 37.79\% & 68.45\% & 88.63 \\
    \ibet{} (Top-5 Langs) & 709.90B & 37.24\% & 68.19\% & 89.24
  \end{tabular}
  }
  \captionsetup{width=0.8\linewidth}
  \caption{Character and n-gram statistics between IB-HL and this dataset (\ibet{}). The \emph{Top-5 Langs} are
    separated because IB-HL performed additional text analysis and OCR cleaning on
  these languages.}\label{table:ib1_v_ibet_1}
\end{table}

\begin{table}[htb!]
\centering
\small
\setlength{\tabcolsep}{5pt}
\begin{tabular}{@{}l r r r r c c@{}}
& & \multicolumn{3}{c}{\textbf{Tokenizability}} & \textbf{Chars (B)} & \textbf{Bigram uniq \%} \\
\cmidrule(lr){3-5}\cmidrule(lr){6-6}\cmidrule(lr){7-7}
  \textbf{Lang} & \textbf{Volumes} & Raw & IB-HL & \ibet{} & IB-HL \hfill \ibet{} & IB-HL \hfill \ibet{} \\
\midrule
English & \numprint{487342} & 90.05 & 97.02 & \textbf{97.65} & 415.4 \hfill 402.6 & 33.66 \hfill 33.02 \\
German & \numprint{157754} & 71.41 & 75.95 & \textbf{76.44} & 147.2  \hfill 143.8 & 46.80 \hfill 46.49 \\
French & \numprint{135871} & 75.15 & 79.68 & \textbf{80.34} & 115.3  \hfill 112.8 & 39.86 \hfill 39.30 \\
Italian & \numprint{46074}  & 71.16 & 74.59 & \textbf{75.15} & 32.1  \hfill 31.5   & 47.25 \hfill 46.81 \\
Spanish & \numprint{28836}  & 76.88 & 80.89 & \textbf{81.69} & 19.6  \hfill 19.2   & 39.45 \hfill 38.62 \\
\end{tabular}
  \captionsetup{width=0.8\linewidth}
  \caption{Per-language comparison on the \numprint{855877} volumes with IB-HL post-processing.
  Excludes 34 volumes without post-processing in IB-HL\@.}\label{table:ib1_v_ibet_2}
\end{table}

\begin{table}[htb!]
\centering
\small
\setlength{\tabcolsep}{5pt}
\begin{tabular}{@{}l c c c@{}}
  \textbf{Group} & \textbf{Volumes} & \textbf{\shortstack{IB-HL \\ Tokenizability}} & \textbf{\shortstack{\ibet{} \\Tokenizability}} \\
  \midrule
  Top-5 Langs & \numprint{855911} & 88.63 & 89.24 \\
  Not Top-5 Langs & \numprint{127091} & 64.56 & 68.53 \\
  \textsc{Nupunkt} Langs & \numprint{962373} & 85.85 & 86.82 \\
  \tallparens{\centstack{\textsc{Nupunkt} Langs \\ Excl. Top-5 Langs}} & \numprint{106462} & 63.53 & 67.33 \\
  \textsc{SaT} Langs & \numprint{20629} & 69.84 & 74.72
\end{tabular}
  \captionsetup{width=0.8\linewidth}
  \caption{Per-group comparison of tokenizability between IB-HL and \ibet{}.}\label{table:ib1_v_ibet_3}
\end{table}

IB-HL applied additional processing on books in the top 5 languages.\footnote{IB-HL
  omitted 34 books having one of the 5 most common primary languages from additional processing.
This causes some small differences in tables in this section.}
The second set of rows in the table restricts to volumes with additional processing in IB-HL and
shows small decreases in bigram and trigram unique ratios and a small increase in tokenizability.
Table~\ref{table:ib1_v_ibet_2} contains a per-language breakdown for the top 5 languages and
demonstrates small, but consistent, differences.
Table~\ref{table:ib1_v_ibet_3} shows that the largest difference in tokenizability comes from
volumes in \textsc{SaT} languages.

Analysis shows that this pipeline increased tokenizability for every language. The largest
increases are Occitan (+9.5), Persian (+5.8), and Arabic (+5.6).

\subsection{Multilingual Reuse of Duplicated Text}\label{ssec:multilingual_reuse}
Corpus-wide duplicate identification and language detection allow tracking how often a passage in
one language occurs \emph{outside} books in its language.
For every duplicated subtopic paragraph, we compare its detected language against the
primary language of the volume that contains it.
Across all \numprint{124049981} duplicated paragraphs, only $5.3\%$ have a detected language
different from their host volume.

\begin{table}[htb!]
  \centering
  \small
  \setlength{\tabcolsep}{6pt}
  \begin{tabular}{@{}l r c c l@{}}
    \toprule
    \textbf{\shortstack[l]{Paragraph\\language}} & \textbf{\shortstack[r]{Dup.\ pars\\($\geq$200 ch)}}
      & \textbf{\shortstack{Own-lang \%\\(all lengths)}} & \textbf{\shortstack{Own-lang \%\\($\geq$200 ch)}}
      & \textbf{Top host languages ($\geq$200 ch)} \\
    \midrule
    English    & \numprint{76448359} & 98 & 99 & German, French ($<$1\% each) \\
    German     & \numprint{7240400}  & 94 & 97 & English 2\%, French 1\% \\
    Russian    & \numprint{144414}   & 96 & 97 & German 1\%, Bulgarian 1\% \\
    French     & \numprint{7862913}  & 93 & 95 & German 2\%, English 2\% \\
    Dutch      & \numprint{275237}   & 81 & 93 & German 2\%, French 2\% \\
    Spanish    & \numprint{519462}   & 88 & 92 & English 6\%, French 1\% \\
    Italian    & \numprint{812897}   & 85 & 91 & German 4\%, English 3\% \\
    Danish     & \numprint{152055}   & 44 & 87 & English 4\%, German 3\% \\
    Portuguese & \numprint{87953}    & 57 & 82 & English 7\%, Spanish 6\% \\
    Latin      & \numprint{1466807}  & 53 & 61 & English 15\%, German 10\% \\
    Greek      & \numprint{82451}    & 51 & 55 & English 25\%, German 10\% \\
    \bottomrule
  \end{tabular}
  \captionsetup{width=0.9\linewidth}
  \caption{How often duplicated paragraphs appear outside volumes of their own
    language. For each detected paragraph language we give the share of its
    duplicated paragraphs hosted in a same-language volume, computed over all
    duplicated paragraphs and over only those of at least $200$ characters.
    Host-language shares and counts are for the $\geq$200-character
  subset.}\label{table:dup_crosslingual}
\end{table}

Some of this cross-lingual sharing is an artifact of language detection on short fragments.
Roman-numeral lists, headers of tables, marginalia-based citations, and other 
commonly mis-OCRed text confuse our language detection algorithms.
Paragraphs of at least 200 characters are predominantly prose, and the added length gives enough
context for more accurate language determination.
In Table~\ref{table:dup_crosslingual}, we describe the $2.3\%$ of
paragraphs having at least $200$ characters and that have a detected language that differs from
their host volume.

The largest languages in \ibet{} are strongly same-language-bound: 95--99\% of duplicated English,
German, French, and Russian ($\ge200$ ch) paragraphs occur in volumes whose primary language
matches. Classical and scholarly languages behave differently.
Latin and Greek paragraphs often appear in texts with different primary languages.
This distribution reveals the collecting practices of Harvard Library as much as
it reveals trends in reusing text. The reuse patterns are consistent with text circulation: legal,
classical, and liturgical passages are quoted verbatim across scholarly literature written in many
languages.

The difference between the distributions among all paragraphs and $\ge200$ ch paragraphs shows
biases within our language detection algorithm. Disproportionately many small subtopic paragraphs
are recognized as Danish or Portuguese. The relative consistency of Latin and Greek across the
two distributions suggests true reproduction and quotation instead of noise.

\subsection{Bits-Per-Byte Over Time}\label{ssec:bpb_time}
\begin{figure}[htb!]
  \centering
  \includegraphics[width=\linewidth]{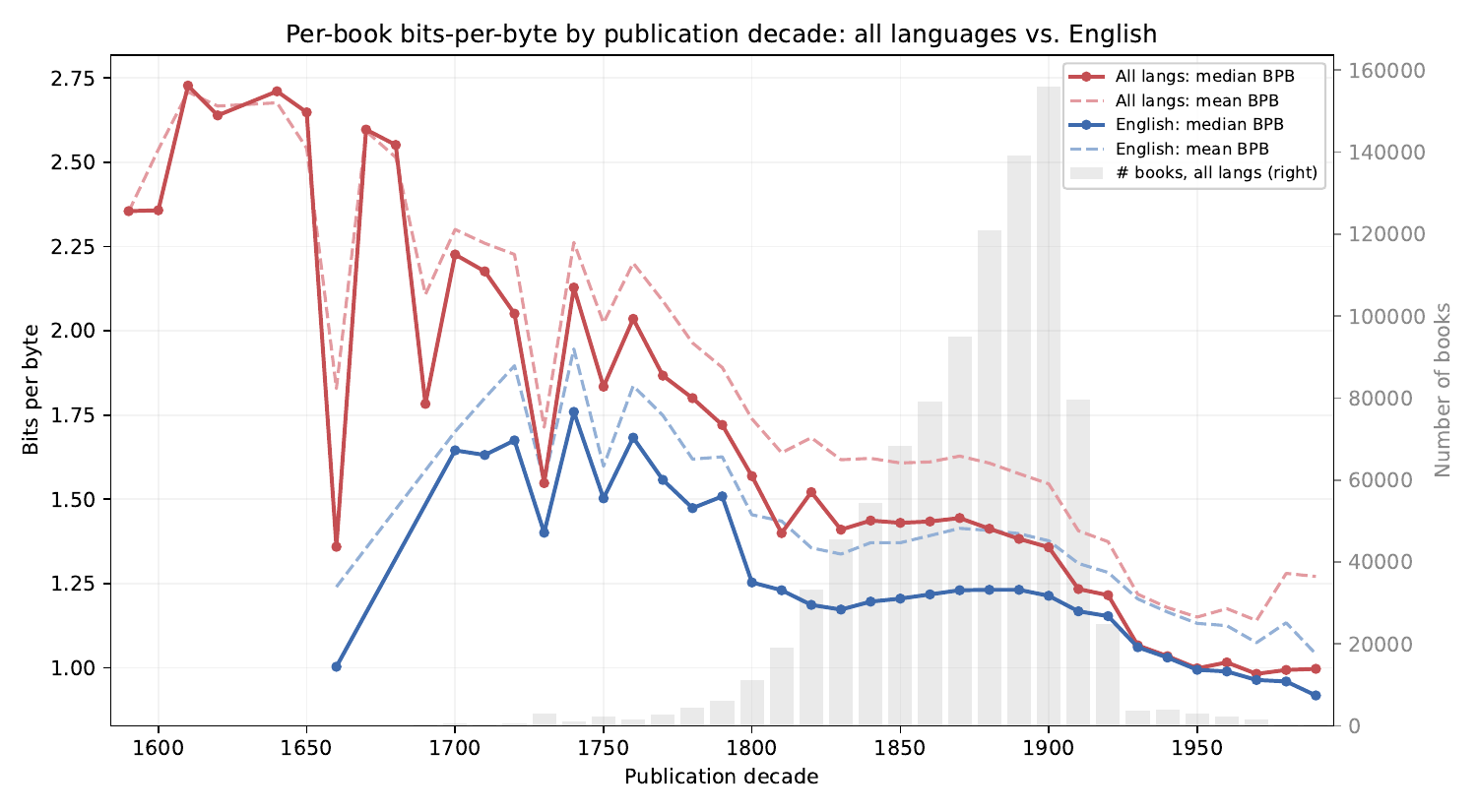}
  \captionsetup{width=0.9\linewidth}
  \caption{Per-book bits-per-byte by publication decade, all languages (red) versus
    English (blue). Solid lines are the median of per-book median BPB\@; dashed lines
    the mean of per-book mean BPB\@. Grey bars give the number of dated books per
    decade (right axis). Decades with fewer than $50$ books are omitted.}%
  \label{figure:bpb_by_year}
\end{figure}

IB-HL compiled publication years for every volume. In this pipeline, we computed the
BPB of every subtopic paragraph. Combining these two distributions allows comparing BPB
across centuries of printing (see Figure~\ref{figure:bpb_by_year}). The dominant feature is a
gradual decline: text becomes more predictable to \textsc{Qwen3-0.6B-Base} as it approaches the
present. Median per-volume BPB falls from $\approx2.5$ in the 17th century to $\approx1.4$ in the
mid-19th century and $\approx1.0$ by the mid-20th century.

We compare English (blue) against all languages (red). The English
subset traces approximately the same shape\footnote{The 1660s have an unusually low BPB\@. This is
due to 58 volumes of \emph{Acts and Resolves passed by the General Court} (the colonial
Massachusetts Bay legislative record). These have high OCR quality, have formulaic statutory prose,
and have correspondingly low median BPB. The average BPB among the 21 other English volumes from
that decade is $\approx1.64$.} but with a slightly smaller median.

One interpretation of this trend is that BPB measures surprise under a modern language model, hence
this curve approximates \emph{distance to contemporary prose}. Older orthography, vocabulary,
typesetting, and styles are less predictable to modern language models. The larger amount of
available text from the 19th and 20th centuries (corresponding to the increased number of available
volumes from that era) might appear in model training data, consistent with the observed decrease in
BPB\@.

We note two potential confounding variables: OCR noise and sample composition.
Due to physical degradation and different publishing norms, older volumes have less consistent
OCR\@. BPB conflates model surprise with OCR noise, hence older volumes likely have higher BPB
purely for OCR reasons.\footnote{ On the other hand, OCR confidence metadata indicates that English
OCR is relatively consistent, and hence the English curve suggests a clear trend and not merely OCR
noise. Disentangling these variations requires additional research. } In addition, the number
of available volumes drops sharply at the public-domain boundary in the 1930s. The available volumes
from after this cutoff are not representative of all that was printed. Continued BPB decline after
1930 may be partially explained by selectivity bias among volumes that already appear in the public
domain.

\section{Discussion and Further Directions}

Like the dataset on which it is based, \ibet{} is a step in an ongoing, collaborative process rather
than a finished artifact. By choosing to \emph{annotate} rather than \emph{remove} paratextual
elements and metadata, we hope this dataset will be a useful starting point that others can easily
tailor to their needs. The philosophy of annotation over removal invites extension: as additional
use cases and needs arise, we may identify additional annotations or analyses to include.
In time, we hope this methodology and the resulting data format will prove
useful to others in creating datasets from cultural texts.
We believe that the adoption of a standardized format for enriching text has the
potential to make data practices in AI model training more robust while
improving the overall computational addressability and cross-compatibility of
text corpora.

\paragraph{Analysis of the Enriched Text.}
Beyond serving as an input to model training, the enriched dataset opens lines of
research that a flattened token stream would not. Earlier analyses (cf.\ \S\ref{sec:analysis})
illustrated a few.
Because duplicate paragraphs are marked rather than deleted
(\S\ref{ssec:deduplication}), the collection is a map of textual reuse:
Table~\ref{table:dup_crosslingual} and \S\ref{ssec:multilingual_reuse} reveal
a rich, cross-lingual structure of quotation and clustering.
Pairing per-paragraph BPB with publication year (Figure~\ref{figure:bpb_by_year},
\S\ref{ssec:bpb_time}) demonstrates that the corpus enables study of historical language change,
raising questions about per-language writing trends, era-specific OCR degradation, and training data
availability. We hope that \ibet{} enables many research directions, including directions
we have not anticipated.

\paragraph{Pipeline Improvements.}
Limitations in  recognizing tables, marginalia, and technical language lead to weaknesses in
the current dataset.
Examples of two known weaknesses are shown in Appendix~\ref{appendix:s:bad}.
We expect advances in OCR will address each of these, in turn improving the
pipeline.

This pipeline was built against the OCR provided for Harvard Library's
collection. It is possible that different collections and different OCR
pipelines may reveal different failure modes and pose new engineering challenges.

\paragraph{Towards a Virtuous Cycle.}
We are working to establish diverse collaborations on this and related datasets. We want this work,
together with IB-HL, to make millions more books computationally accessible.
We also hope that improved training data for low-resource languages will lead to improved models,
allowing more accurate analysis across the collection.
As noted by~\cite{beyene2026survey}, there is a growing body of work demonstrating technical
capacity to process historical documents, but a lack of prioritization in training data and
evaluation benchmarks.
We release the pipeline and data with the goal that they are extended and applied to further
collections. As communities extend the annotations, develop new cleaning techniques, correct errors,
and add perspectives, those improvements can be incorporated into the source material.
Crucially, that cycle must preserve provenance, reproducibility, and the ability to audit
transformations.

\section{Acknowledgements}

We thank Harvard Library for their support and for giving us the opportunity to
continue working on this unique collection.
Many computations in this paper were run on the FASRC Cannon cluster supported by the FAS Division
of Science Research Computing Group at Harvard University.
We thank Kacie Bailey, Matte Hartog, and Jimmy Mendez for their support and encouragement.

We also thank Ted Underwood, Nick Levine, and Alec Radford for helpful comments and suggestions,
especially regarding annotation, normalization, and bits-per-byte filtering. Incorporating this
feedback has led to an improved pipeline, dataset, and report.

The pipeline uses many libraries and projects mentioned in this report. We also credit technical
libraries that we used for exploration, data analysis, prototyping,
distributing work across compute nodes, and preparing this technical report. We used
\textsc{Jupyter}~\parencite{jupyter}, \textsc{Pandas}~\parencite{pandas_paper, pandas_software},
Apache \textsc{Arrow}, and \textsc{scikit-learn}~\parencite{scikitlearn, scikitlearn_api} in
essential ways at many stages of exploration and development. We used \textsc{pybind11}~\parencite{pybind11}
to write Python extensions in C++ for performance, orchestrated tests using
\textsc{Pytest}~\parencite{pytest} to ensure stability across heterogeneous compute environments,
distributed runs using \textsc{GNU Parallel}~\parencite{gnu_parallel}, and monitored progress using
\textsc{tqdm}~\parencite{tqdm} bars and \textsc{loguru} logging. Finally, we used
\textsc{NumPy}~\parencite{numpy} and \textsc{matplotlib}~\parencite{matplotlib} to generate all
figures in this report.

This work was supported by unrestricted funding from Microsoft, OpenAI, Meta, and Jane
Street.

AI tools were used to assist in the preparation of this technical report.

\section{Rights Determination}%

We respect the intellectual property rights of authors, publishers, and other rights holders. While
we have taken deliberate steps to include only those volumes for which there is no known copyright
restriction, specifically those identified by the HathiTrust Digital Library with a status of
``public domain,'' ``public domain in the United States,'' or ``CC-Zero,'' copyright determinations are
complex and context-dependent, and occasionally subject to error.

While this is relatively low risk, some volumes in this dataset may be in the public domain in the
United States but still subject to copyright or other rights protections in other jurisdictions.
Additionally, the absence of an explicit copyright claim or rights status does not guarantee that a
work is in the public domain, either in the U.S. or abroad. Information about the copyright status
of individual volumes is provided on a good-faith basis and reflects available data at the time of
determination, but we cannot guarantee its completeness or accuracy.

Users of this dataset will be solely responsible for making independent legal assessments about how
and where they use the materials. Some uses of materials may also be restricted by trademark,
privacy, publicity rights, or other such rights or restrictions. It is the user's sole
responsibility to consider the possibility that such rights or restrictions may be involved and to
secure any needed permissions. If any rights holder believes that a work included in this release is
misidentified or improperly included, we welcome contact and will promptly review any concerns. Our
goal is to provide broad public access while maintaining respect for intellectual property rights
and ensuring responsible data stewardship.

\section{Disclaimers}
\label{sec:disclaimers}

\refstepcounter{subsection}
\subsection*{\thesubsection\quad Harmful Language and Content in this Dataset}
This dataset is a collection of historical works that reflect the language, imagery, culture, and
perspectives of their time. Users should be aware that some materials may contain language or
portrayals that are outdated, offensive, or harmful today, such as racism, sexism, colonial
attitudes, and other forms of discrimination. Some content may include inaccurate information,
providing insight into historical contexts that existed at the time of writing. The materials are
maintained in their original form to retain contextual understanding and facilitate research
efforts, but we encourage critical awareness and cultural sensitivity for the creators and/or
subjects of the collection. These materials are offered as part of a historical perspective, but
should not be considered a stand-alone research collection constructed to give a balanced
perspective on any topic.

\refstepcounter{subsection}
\subsection*{\thesubsection\quad Harmful Language in Bibliographic Description}

Metadata for this collection may contain language that is overtly or implicitly harmful, outdated,
or biased, or may by omission fail to represent important perspectives. Metadata may contain
language created decades ago. It is common practice within the field of library science to reuse
descriptions provided from the creator of the materials. While in some instances this allows
communities and individuals to represent their materials in their own words, unexamined use of this
practice may mean that racist or other offensive terminologies appear in our description. We also
use national standardized terms in our work that can be outdated and harmful. Note that terminology
in historical materials and in library descriptions does not always match the language we currently
understand to be preferred by members of the communities depicted.

Furthermore, we acknowledge that the act of collecting materials is not always neutral, and the work
of describing and classifying library materials is influenced by inherent personal, institutional,
and societal biases. Outdated or offensive terminologies may be present in metadata such as subject
headings, and harmful language or bias may be introduced by catalogers supplying titles and
descriptions. In other cases, books themselves present racist, offensive or otherwise harmful
viewpoints in titles or descriptions that are routinely transcribed by catalogers. 

Note: Some language in this statement was adopted from Harvard Library's statement on Harmful
Language in Library
collections.\footnote{\url{https://library.harvard.edu/harmful-language-library-collections }}

\refstepcounter{subsection}
\subsection*{\thesubsection\quad Generated and Experimental Content}

This dataset contains generated and/or experimental content. While reasonable
care was taken to ensure its quality, it is provided ``as is,'' without
warranties of any kind. It may contain errors or inaccuracies; users should
verify the data independently and apply their own judgement.

\addcontentsline{toc}{section}{Reference list}
\printbibliography[title=Reference list]{}

\clearpage{}

\section*{Appendices}

\begin{appendices}

\section{Dataset Fields}\label{appendix:s:dataset_fields}

\begin{table}[tbh!]
\centering
\footnotesize
\setlength{\tabcolsep}{4pt}
\renewcommand{\arraystretch}{1.15}
\begin{tabular}{@{}l l p{6.2cm} l@{}}
\toprule
\textbf{Field name} & \textbf{Type} & \textbf{Description} & \textbf{Section} \\
\midrule
\texttt{barcode\_src} & string & The volume's barcode. Serves as a primary key/identifier and join key to IB-HL. & -- \\
\texttt{primary\_language\_gen} & string & ISO 639-3 code for the main language of this volume, taken from IB-HL for convenience. & -- \\
\texttt{language\_distribution\_gen} & list[$\langle$string, float$\rangle$] & Pairs of $\langle$\texttt{LANG}, \texttt{PROPORTION}$\rangle$ where 
        \texttt{LANG} is an ISO 639-3 code and \texttt{PROPORTION} is the $0$--$1$ fraction of the paragraphs in the volume with that language.
        Only languages with $\geq 5$ paragraphs contribute. & \S\ref{ssec:annotation} \\
\texttt{token\_count\_gen} & int & Total tokens for that volume's middlematter, as measured with \textsc{o200k\_base}. & \S\ref{ssec:annotation} \\
\texttt{char\_count\_gen} & int & Total characters in middlematter. & \S\ref{ssec:annotation} \\
\texttt{word\_count\_gen} & int & Total words in middlematter (language-aware tokenization). & \S\ref{ssec:annotation} \\
\texttt{sentence\_count\_gen} & int & Total sentences in middlematter (language-aware segmentation). & \S\ref{ssec:segment} \\
\texttt{paragraph\_count\_gen} & int & Total subtopic paragraphs in middlematter. & \S\ref{ssec:chunk} \\
\texttt{section\_count\_gen} & int & Total subtopic sections in middlematter. & \S\ref{ssec:chunk} \\
\texttt{bigram\_count\_gen} & int & Total word bigrams in middlematter. & \S\ref{ssec:annotation} \\
\texttt{bigram\_count\_unique\_gen} & int & Distinct word bigrams in middlematter. & \S\ref{ssec:annotation} \\
\texttt{trigram\_count\_gen} & int & Total word trigrams in middlematter. & \S\ref{ssec:annotation} \\
\texttt{trigram\_count\_unique\_gen} & int & Distinct word trigrams in middlematter. & \S\ref{ssec:annotation} \\
\texttt{tokenizability\_ratio\_gen} & float (0.0--100.0) & Measure of how close this text is to 1.25 \textsc{o200k\_base} tokens per word. & \S\ref{ssec:annotation} \\
\texttt{bpb\_\{min,max,median,avg\}\_gen} & float & Per-volume bits-per-byte summary over scored paragraphs. & \S\ref{ssec:bpb} \\
\texttt{bpb\_p\{10,30,70,90\}\_gen} & float & Per-volume bits-per-byte percentiles. & \S\ref{ssec:bpb} \\
\texttt{frontmatter\_gen} & string & Annotated frontmatter: contains endmatter divs. & \S\ref{ssec:endmatter} and \S\ref{ssec:annotation} \\
\texttt{middlematter\_gen} & string & Annotated body: contains sections, paragraphs, and bpb/language/duplicate tags. & \S\ref{ssec:annotation} \\
\texttt{backmatter\_gen} & string & Annotated backmatter: contains endmatter divs. & \S\ref{ssec:endmatter} and \S\ref{ssec:annotation} \\
\texttt{processed\_middlematter\_gen} & string & Middlematter filtered to non-duplicate paragraphs in the $[\mathrm{p}10,\mathrm{p}90]$ bpb band. & \S\ref{ssec:annotation} \\
\bottomrule
\end{tabular}
  \captionsetup{width=0.8\linewidth}
  \caption{Fields in the dataset.\newline The suffix \texttt{\_src} denotes fields
    taken directly from source inputs. The suffix \texttt{\_gen} denotes values
    computed by this pipeline. All text statistics are computed over
    middlematter.}\label{tab:dataset-fields}
\end{table}

\section{Additional related work}\label{appendix:related}

Our work incorporates many different active areas of research and engineering. We indicate relevant
work, especially work that informed our design, even if we do not use it directly.

\begin{sloppypar}
\paragraph{Cleaning Datasets.}
Computationally inexpensive boilerplate detection is well studied~\parencite{weninger2010cetr,
wang2015qread, kohlschutter2010boilerplate}.
OCR post-analysis and detection is its own field of research, with some methods based on statistical
analysis~\parencite{kukich1992techniques, tong1996statistical,
jatowt2019deep}.
See~\cite{nguyen2021survey} for a survey describing different approaches.
More recently, modern ML-based techniques (especially models based on transformers) have been used.
ICDAR held competitions in 2017 and 2019 on OCR correction~\parencite{chiron2017icdar2017,
rigaud2019icdar2019} in which neural networks and BERT-based models~\parencite{devlin2019bert}
outperformed most other methods.
\end{sloppypar}

\paragraph{Data Curation.}
Assembling public datasets for model training is a multistage process that can include data
sourcing, deduplication, cleaning, and filtering.
The C4 dataset~\parencite{raffel2020exploring} is a large-scale
dataset from internet-scraped data, processed with heuristic cleaning, English language restriction,
and chunk-based deduplication.

One difficulty in this area is that details about the training process and data for state-of-the-art
language models are often opaque.
Even open-weight models such as LLaMA, Mistral, Gemma, GLM, and Kimi reveal
little about underlying training data~\parencite{llama, mistral7b, gemma,
glm5, kimi_k2}.
To improve transparency and to facilitate comparison, DCLM~\parencite{li2024datacomp}
establishes benchmarks for data curation.

\paragraph{Landscape Surveys.}
The progression from large data curation~\parencite{gao2020pile,
raffel2020exploring}, to filtering and extraction~\parencite{wenzek2020ccnet}, to post-creation
analysis~\parencite{dodge2021documenting} helped shape our approach to data.
Datacomp-LM~\parencite{li2024datacomp} also provides a thorough literature review and survey of
closely related areas.

\section{Details on Processing}\label{appendix:details}

This appendix contains additional technical details and design rationale.

\subsection{Preprocessing Details}\label{appendix:ssec:preprocessing}

\paragraph{Language-based Models.}
This setup relies on language reporting from IB-HL, as we use that language detection for
stratification.
Each of the models here is a \emph{base} model and is later updated for each book processed.
A small amount of noise in the base model has minimal effect on the resulting
behavior in our pipeline.
An additional source of noise comes from training base models on pre-cleaned text. For most
languages and most volumes, however, the majority of the text has sufficiently little noise to allow
effective use of small base-language models.

\paragraph{Segmenter-based Partitioning.}

The IB-HL dataset can be naturally streamed in shards of $200$ books. Rather than use those shards, we
immediately split shards into \textsc{Nupunkt}-compatible languages and non-\textsc{Nupunkt}-compatible languages. We
do this because \textsc{Nupunkt}-compatible shards can be processed entirely on CPUs, while the other shards
require GPUs to complete sentence segmentation in a reasonable amount of time. After this sharding,
all further processing (except deduplication) is completely independent.

\subsection{Unicode Normalization}\label{appendix:sss:unicode}

\paragraph{Soft and Hard Normalization.} See Listing~\ref{listing:unicode} for the exact steps for both ``soft'' and ``hard'' normalization.
We apply NFC and NFKC normalization via \texttt{unicodedata.normalize} in the
Python standard library.
The remaining steps are strict string replacement and regular expression operations.

\begin{listing}[htb!]
  \begin{minipage}{0.45\linewidth}
    {\footnotesize
      \begin{minted}[frame=single, framesep=2mm]{text}
Soft Unicode Normalization

1. NFC Unicode Normalization
2. Remove zero-width characters:
   U+200B, U+FEFF
3. Condense Unicode space characters:
   Map U+00A0, U+2000-U+200A, U+202F,
   U+205F, and U+3000 to ASCII space
4. Collapse consecutive spaces
5. Strip per-line leading/trailing spaces

PRESERVES: line breaks, curly quotes,
hyphens, dashes, ligatures, accents.
      \end{minted}
    }
  \end{minipage}
  \hfill{}%
  \begin{minipage}{0.50\linewidth}
    {\footnotesize
      \begin{minted}[frame=single, framesep=2mm]{text}
Hard Unicode Normalization

1.  NFKC Unicode Normalization
2.  Remove zero-width characters:
    U+200B, U+FEFF
3.  Map newlines '\r\n' and '\r' to '\n'
4.  Map hyphen/dash-like to ASCII dash
    U+2010-U+2015, U+2212, U+FE58, U+FE63,
    U+FF0D to ASCII dash.
5.  Remove U+00AD soft-hyphen.
6.  Map curly quotes to ASCII quote
    U+2018-U+201B and backtick to '
    U+201C-U+201F to "
7.  Map Unicode space characters to ASCII space
    U+00A0, U+2000-U+200A, U+202F,
    U+205F, and U+3000
8.  Flatten tabs and newlines to ASCII space
9.  Collapse consecutive spaces
10. Strip per-line leading/trailing spaces

OUTPUT: a single line with ASCII punctuation
and NFKC-folded characters.
      \end{minted}
    }
  \end{minipage}
  \caption{Explicit Unicode Normalization}\label{listing:unicode}
\end{listing}

\begin{table}[htb!]
  \centering
  \small
  \begin{tabular}{@{}l l c c@{}}
    \toprule
    \textbf{Case} & \textbf{Input} & \textbf{Soft} & \textbf{Hard} \\
    \midrule
    Precomposed vs.\ combining accent & \texttt{e}~+~\texttt{U+0301} & \'{e}~(NFC) & \'{e} \\
    Ligature                          & \texttt{U+FB01}~(\,ﬁ\,)       & ﬁ~(kept)    & \texttt{fi} \\
    Full-width letter                 & \texttt{U+FF21}~(\,\supp{Ａ}\,)      & \supp{Ａ}~(kept)    & \texttt{A} \\
    Superscript digit                 & \texttt{U+00B2}~(\,\textsuperscript{2}\,) & \textsuperscript{2}~(kept) & \texttt{2} \\
    Curly quotation marks             & \texttt{U+201C}\,\ldots\,\texttt{U+201D} & ``\,\ldots\,'' & \texttt{"}\,\ldots\,\texttt{"} \\
    En / em dash, minus sign          & \texttt{U+2013 / U+2014 / U+2212} & --\,/\,---\,/\,$-$ & \texttt{-} \\
    Non-breaking space                & \texttt{U+00A0}                & ASCII space & ASCII space \\
    Zero-width space                  & \texttt{U+200B}                & (removed)   & (removed) \\
    \bottomrule
  \end{tabular}
  \captionsetup{width=0.85\linewidth}
  \caption{Soft vs.\ hard Unicode normalization on boundary cases.}\label{table:unicode_examples}
\end{table}

\paragraph{Zero-Width Spaces.}
One aspect of zero-width spaces merits additional discussion. The zero-width spaces U+200C
(zero-width no-join) and U+200D (zero-width yes-join) are kept, despite not being visible. These are
mainly used in cursive Arabic, Persian, and Indic scripts to either prevent characters joining in
cursive or to force joining in cursive. These zero-width spaces can help distinguish word or
morpheme boundaries and carry semantic meaning. (U+200D is also used in some emoji sequences, but
this is not relevant for this corpus). As OCR continues to improve for cursive Arabic, Persian, and
Indic scripts, these zero-width formatting spaces will likely become more prevalent.

\paragraph{Boundary Examples.}
Table~\ref{table:unicode_examples} contrasts the ``soft'' and ``hard'' normalization on
representative inputs.

\subsection{Duplicate Page Removal}\label{appendix:ss:duplicate_page}

For duplicate page removal, we first hard-normalize each page in a volume (cf.\
\S\ref{ssec:unicode}). Thus each page is temporarily reduced to a single long string of text for
duplicate comparison.

\paragraph{Minimum Page Length.}
To ensure that pages have sufficient text to be worth checking for duplicates,
we only consider pages with at least $50$ non-whitespace Unicode code points (after
hard-normalization).
Technically this carries a small bias against detecting duplicates in languages
that have more semantic meaning per character. For example, $50$ Chinese characters
likely carry more semantic content than $50$ ASCII characters.
This is a mild lower bound that prevents excessive removal of short content.

\paragraph{N-Gram Length.}
The simhashes are computed over character 9-grams. Longer n-grams make the hash more sensitive to
exact content and less prone to random short common phrase collision. Shorter n-grams allow more
n-grams and random mixing per page. We chose 9 to balance these competing requirements.

\paragraph{Compiled C++ Simhash Extension.}
We use the MurmurHash3 non-cryptographic hash function for rapid hashing; it is faster than
cryptographic hash functions such as SHA256, and cryptographic security is not required.
MurmurHash3 was specifically designed for high performance after compilation~\parencite{smhasher}.
The standard implementation for Python, \\ \textsc{mmh3}~\parencite{mmh3_murmurhash} is written in C
and computes hashes quickly, but passing Python objects to \textsc{mmh3} and manipulating the
hashes in Python adds overhead.

We include a lightly customized version of MurmurHash3 in an optimized C++ simhash implementation.
This implementation is designed for 64-bit platforms (little-endian only), uses GCC/Clang optimized
\textsc{popcnt} for rapid Hamming distance calculation between simhashes, and does not need to pass
objects back to Python for comparison.
Benchmarks show that this results in an approximately 100-fold speedup over a direct Python
implementation of simhash using \textsc{mmh3}.
We verify that our MurmurHash3 implementation agrees with \textsc{mmh3} in a test suite.

\paragraph{Probability Analysis.}

Page simhashes that differ by 6 or fewer bits are recognized as duplicates.
This is a strict threshold and should rarely occur by chance. In detail: for any fixed
128-bit integer, the number of other 128-bit integers with Hamming distance at most $6$ is
\begin{equation*}
  B(128, 6) = \sum_{k = 0}^6 \binom{128}{k} \approx 5.70 \times 10^9.
\end{equation*}
Thus the probability that two random hashes will have Hamming distance at most
$6$ is approximately
\begin{equation*}
  B(128, 6) / 2^{128} \approx 1.67 \times 10^{-29}.
\end{equation*}
For a collection of $N$ hashes, the expected number of accidental 6-bit near-collisions is
approximately $\dbinom{N}{2} \cdot 1.67 \cdot 10^{-29}$. Hence even for the largest books,
the expected number of accidental collisions is negligible.

\subsection{Endmatter Separation}

\paragraph{Compute Parameters.}
Pages are batched, but at most from a single book at a time. We found empirically that batches of
size $\approx1024$ are optimal (capable of 7200 pages/s per CPU core). Book page lengths resulted in
our typical batch size being closer to 350 (6800 pages/s per CPU core). In practice, total memory
bandwidth is a limiting factor and using 20 cores is not 20 times faster than using 1 core.

\begin{sloppypar}
\paragraph{Bottleneck.}
Benchmarks show that the embedding table of dimension $512$ limits overall speed.
The embedding table used by \textsc{Model2Vec} has a 250k vocabulary (inherited from
\textsc{BAAI/BGE-M3}), each mapping to $512$ dimensions with a 4-byte \texttt{float32}.
Hence the embedding table is approximately 512MB and does not fit into cache on most CPUs.
Most table access requires memory lookup.
\end{sloppypar}

\paragraph{Multilingual Generalization.}
We apply the same models to every book in every language.
We believe that the multilingual training supporting \textsc{BAAI/BGE-M3} helps
provide good signal across the languages in IB-HL. For languages not covered in its training data,
one can sometimes expect a small degree of zero-shot generalization for languages that share scripts
or lexical features with those seen during training~\parencite{bgem3}.

\paragraph{Generalized Training Data.}
The pipeline repository includes code that trains analogous classifiers based on given
data. The repository allows configuring static models for
\textsc{Model2Vec}, including the option to bypass static models entirely.
The repository is designed to facilitate training classifiers on alternative data.

\begin{listing}[p!]
  \centering
  {\footnotesize
\begin{minted}[frame=single,framesep=2mm]{python}
  SYSTEM_PROMPT = """
  You create realistic book content pages that reflect both historical and linguistic
  accuracy. Your task is to generate sample pages in a specified language, adhering to the
  conventions of the given era.
  """
  GOOD_PAGE_TYPES = [ "Fiction Novel", "Non-fiction", "Textbook Exposition",
    "Argumentative Essay", "Documentation from a how-to manual",
    "Dialogue from a play or interview", "Travel Writing", "Academic Writing",
    "Letters of Correspondence", "Poetry" ]
  BAD_PAGE_TYPES = [ "Table of Contents", "Copyright", "Dedication Page", "Index Page",
    "Bibliography", "Section Dividers", "Frontmatter", "Backmatter" ]
  ERAS = [ "pre1800s", "1800-1825", "1826-1850", "1851-1875",
     "1876-1900", "1901-1925", "modern", ]
  GOOD_PROMPT = """
  You will generate pages from TWO fictional books in the language '{language}' (ISO 639-3
  code) set in the {era} era, but with contrasting content.

  For EACH of the two books, generate ONE COMPLETE EXAMPLE PAGE found in a {page_type}
  type of book.

  Output requirements:

  * Total pages: TWO (one for each book).
  * After each page, insert a line with exactly: #####
  * The content of each page must be in the target language only.
  * Make the tone and layout appropriate to the genre of each book.
  * Do not include any translations or meta-commentary.
  * If you cannot write in the language for '{language}', output exactly UNSUPPORTED_LANGUAGE
  """
  BAD_PROMPT = """
  You will generate frontmatter and backmatter pages for TWO different fictional books in
  the language '{language}' (ISO 639-3 code), both set in the {era} era, but with
  contrasting genres.

  First book: a formal, scholarly, or religious work.
  Second book: an informal, narrative, or literary work (such as a novel, memoir, or story
  collection).

  For EACH of the two books, generate ONE COMPLETE EXAMPLE PAGE found in the {page_type}
  of the book.

  Output requirements:

  * Total pages: TWO (one for each book).
  * After each page, insert a line with exactly: #####
  * The content of each page must be in the target language only.
  * Make the tone and layout appropriate to the genre of each book.
  * Do not include any translations or meta-commentary.
  * If you cannot write in the language for '{language}', output exactly UNSUPPORTED_LANGUAGE
  """
\end{minted}
}\caption{Synthetic Data Prompt Structure}\label{listing:synthetic}
\end{listing}

\subsection{Dehyphenation}\label{appendix:ss:dehyphenation}


\paragraph{Two Models are Necessary.}
Both the base model and the per-book model are necessary.
Per-book models lack statistics for long words that appear only once in a book.
Long names typically appear in only a few books, and hence are not contained in
base language statistical distributions.
Combining per-book and per-language base statistics leads to more reliable dehyphenation decisions.

\paragraph{Probabilistic Interpretation.}
Our model is a ``naive joint probability estimator with add-$k$ smoothing and fixed
\emph{Stupid Backoff}~\parencite{brants2007large} penalty.''
We call it \emph{naive} because all probabilities are assumed independent.
The model is not a proper probability model because the backoff does not take into account the probability
mass from full n-grams (distinguishing it from more advanced Katz-style
backoff~\parencite{katz1987estimation}).
Though independence is objectively false and the backoff model is simple, these simplifications have
minimal effect on the relative ranking among the three possibilities in practice.
It would not be appropriate, however, to rely on our probability estimator to generate
text.

\paragraph{Alternatives Considered.}
Prototypes used the KenLM~\parencite{kenlm} implementation of a full Kneser-Ney
smoothed n-gram language model~\parencite{kneserney1994ngramlm}. This is a more precise and robust
language model that offers proper conditional probabilities.
However, KenLM was substantially more computationally expensive (adding $\approx10$s per book).
Experimenting with less computationally expensive alternatives showed that our simple joint
probability estimator in nearly all cases generates the same relative ranking at substantially lower
cost.

\subsection{Running Headers and Footers}

We tuned the Jaccard similarity threshold of $0.85$ to allow approximately one typo or one digit to
change in page numbers for a standard, relatively short header. Experiments show that this tends to
be conservative in practice.

We did not optimize this portion of the pipeline.
We use the default hash algorithm from the \textsc{datasketch} library~\parencite{datasketch}, with
default seeding and the SHA-1 hash function.
It would be possible to use a fast hash function like MurmurHash3 (as in
Appendix~\ref{appendix:ss:duplicate_page}), but the bottleneck is in LSH construction.
For each n-gram, \textsc{datasketch} performs a hash \emph{and} a NumPy permutation-and-min
operation. The permutation-and-min dominates timing.

\subsection{Page Number Removal}

Removing page numbers across different writing systems faces many edge cases. We rely on the
Unicode category of the underlying symbol: if the Unicode category starts with ``\texttt{N}'', we
call it numeric. Otherwise it is non-numeric.
This accepts bare numbers (123), numbers with a stray mark (p 42), Arabic-Indic numbers, Devanagari
numbers, full-width Unicode numbers, Unicode encoded Roman numerals (such as U+2163 for \supp{Ⅳ} instead
of ASCII IV), and so on.
It does not accept certain less common page number formats, such as dash-flanked numbers
(\texttt{- 12 -}).

Most current OCR engines do not use Unicode encodings for Roman numerals and instead use ASCII
letters. Another known gap comes from Han ideographic numerals. For example, `\supp{十 百}' are each
classified in Unicode as \texttt{(Letter, other)} characters and \emph{not} as numbers. Treating
these ideographs as numbers (which is what Python's \texttt{str.isnumeric()} does) would lead to
recognizing legitimate lines such as `\supp{十年}' (``ten years'') as likely page numbers.

\paragraph{Likely OCR Risk.}
Several different scripts use nearly identical shapes with different meanings, e.g.\ the English
digit ``\texttt{0}'' vs the Arabic letter ``\textarabic{ه}'' vs the Devanagari digit ``\devazero{}''
vs the Greek letter $\omicron$ vs a small circle $\circ$. Each OCR engine produces its best estimate
from context and runtime configuration.

\subsection{Segmentation Details}\label{appendix:ss:segmentation}

The list of \textsc{Nupunkt}-compatible languages is in Table~\ref{table:nupunkt_languages}.
\begin{table}[ht!]
  \centering
  \begin{tabular}{llllllllll}
    \toprule
    als & arl & arn & ast & bel & bem & bin & bos & bre & cab \\
    cak & cbt & ces & chk & chv & cic & cjk & ckb & cnh & cnr \\
    cof & ctd & cym & dan & deu & dga & ekk & ell & eng & epo \\
    eus & ewe & fao & fat & fij & fin & fkv & fra & fry & gla \\
    gle & glg & glv & gyr & hat & haw & heb & hil & hlt & hns \\
    hrv & hsb & hun & ibo & ido & ijs & ilo & isl & ita & kal \\
    kat & kaz & kir & kjh & kmb & kng & koi & ktu & lat & lin \\
    lit & lld & loz & lua & lug & lun & mad & men & mic & min \\
    mlt & mri & nba & nbl & ndo & niu & njo & nld & nno & nya \\
    nym & nyn & oki & oss & piu & plt & pol & por & pov & ppl \\
    que & qug & rar & roh & ron & rus & sah & sco & slk & slv \\
    sme & snk & spa & srp & suk & sun & sus & swb & swe & swh \\
    tam & tat & tgl & tsn & tso & tuk & tur & ukr & ura & ven \\
    vie & war & xho & yao & ykg & yua & zro & zul \\
    \bottomrule
  \end{tabular}
  \caption{\textsc{Nupunkt}-compatible languages, given as ISO 639-3 strings}\label{table:nupunkt_languages}
\end{table}

\paragraph{Determining \textsc{Nupunkt} vs \textsc{SaT} languages.}

To choose between \textsc{Nupunkt} and \textsc{SaT} models for sentence segmentation, we weighed
accuracy and speed.

\textsc{Nupunkt} examines punctuation and performs well when punctuation use resembles
common European languages. \textsc{SaT} uses deep learning. If speed (or total compute) were not an issue,
then we would use 12-layer \textsc{SaT} for all segmentation; small samples suggest that
\textsc{SaT} has better segmentation in almost all cases. But \textsc{SaT} requires vastly larger
amounts of compute.

We randomly sampled 10 books in each language from the collection (or as many as possible if 10 were
not available) and used both \textsc{Nupunkt} and \textsc{SaT} to segment into sentences. We assumed
that \textsc{sat-3l-sm} was ``correct'' and assessed how accurately \textsc{Nupunkt} predicted the
indices for starts of sentences.
Treating the output of \textsc{SaT} as ``correct'' is only a heuristic.
In the unlikely event that a language is well-segmented by \textsc{Nupunkt} but poorly segmented by
\textsc{SaT}, this heuristic would cause the worse segmenter to be chosen.

\begin{figure}[htb!]
  \centering
  \includegraphics[width=\textwidth]{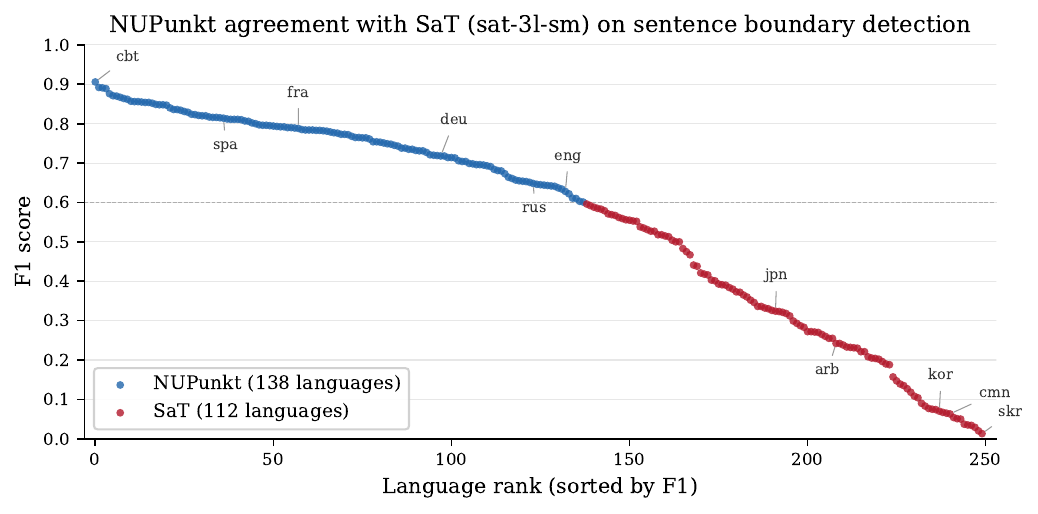} 
  \caption{F1 scores in decreasing order with certain notable languages indicated.}%
  \label{fig:nupunkt_f1}
\end{figure}

A plot of the resulting F1 scores (on identifying \textsc{SaT} sentence start indices) is in
Figure~\ref{fig:nupunkt_f1}.
Instead of a sharp drop, \textsc{Nupunkt} performance gradually decreases.
Benchmarks suggest that both \textsc{Nupunkt} and \textsc{SaT} perform well on
English.
We chose the cutoff between \textsc{Nupunkt} and \textsc{SaT} to be at the
$0.6$ F1 score. English is close to the cutoff, suggesting that \textsc{Nupunkt} should segment all
``\textsc{Nupunkt}-compatible'' languages approximately as well as it segments English.
This cutoff also guaranteed that the vast majority of books in IB-HL would be segmented via
\textsc{Nupunkt}, allowing a predominantly CPU-based pipeline.

\paragraph{\textsc{Nupunkt} per-book adaptation.}

To estimate the value of \textsc{Nupunkt} per-book adaptation, we ran experiments.
We took 195 books (15 books in each of English, French, Portuguese, Spanish, Polish, Russian,
Italian, Latin, Dutch, German, Danish, Greek, and Swedish) and examined concordance between detected
sentence boundaries between each of (1) base language model \textsc{Nupunkt}, (2) per-book adapted
\textsc{Nupunkt},
and (3) \textsc{SaT}. (The setup is similar to the setup leading to Figure~\ref{fig:nupunkt_f1}).

In general, we found that base-language \textsc{Nupunkt} models and per-book adapted
\textsc{Nupunkt} models agree
approximately 93\% of the time, while per-book adapted \textsc{Nupunkt} models agree slightly more with
\textsc{SaT}. Manual checking suggests that the 7\% difference with per-book adapted
\textsc{Nupunkt} models
corresponds to more accurate segmentation.

We conclude that per-book adaptation yields a notable improvement, but may be omitted when compute is
constrained.

\paragraph{More on \textsc{Nupunkt} vs \textsc{SaT} speed comparisons.}

Base-language \textsc{Nupunkt} models perform about twice as fast as \textsc{SaT} in our benchmarks.
Further, on machines with multiple CPU cores, multiple \textsc{Nupunkt} models can operate in
parallel.
We observed that one can run up to 8 simultaneous \textsc{Nupunkt} inference processes on a DGX Spark before
confronting memory bandwidth or RAM limitations.
However, including per-book adaptation changes this tradeoff: 8 simultaneous \textsc{Nupunkt}
adaptation-and-inference processes might segment books at comparable speeds to a single reasonably
fast GPU\@.
Choosing which path and model to use thus depends heavily on whether one has access to GPUs.

We estimate that a single NVIDIA A100 could perform \textsc{SaT} segmentation on all of IB-HL in about 1 month.
For comparison, we estimate that a single DGX Spark could perform the current
\textsc{SaT}+\textsc{Nupunkt}
split segmentation in approximately 2 weeks; this decreases to 3 days when omitting per-book
adaptation.

\paragraph{Alternative Segmenters Considered.}

We ran the segmentation-vs-\textsc{SaT} experiment shown in Figure~\ref{fig:nupunkt_f1} with $4$
other libraries in all languages and $1$ additional library on selected languages. We briefly
comment on these.

\begin{enumerate}[noitemsep,nosep]
  \item
    \textsc{mwtokenizer}\footnote{\url{https://gitlab.wikimedia.org/repos/research/wiki-nlp-tools/}},
    a multilingual Python-based tool used widely on Wikipedia. It ran faster than
    \textsc{Nupunkt} and had comparable behavior on Western European languages, but was
    inconsistent  with other languages.

  \item \textsc{sentencex}\footnote{\url{https://github.com/wikimedia/sentencex}}, another
    segmenter used by Wikimedia. It achieved excellent accuracy but was too slow for this application.

  \item
    \textsc{pySBD}\footnote{\url{https://github.com/nipunsadvilkar/pySBD}} ``Pragmatic Sentence
    Boundary Disambiguation''~\parencite{pysbd}. It had strong performance on 15 languages, but
    was over $10\times$ slower than \textsc{Nupunkt}. Experiments suggest that its runtime
    grows super-linearly in text length, complicating efforts to segment book-length volumes.

  \item \textsc{blingfire2}\footnote{\url{https://pypi.org/project/blingfire2/}},
    developed by the Microsoft BLING (Beyond Language understandING) team.
    Performance and speed are comparable to \textsc{Nupunkt}.

  \item \textsc{NLTK libraries}\footnote{\url{https://www.nltk.org}} for the Python natural language
    toolkit. Most languages have at least one high-performing language-specific NLTK
    library, but there are many and we did not allocate time to test hundreds of NLTK
    libraries. The default English segmenter was less reliable than \textsc{Nupunkt}.
\end{enumerate}

One distinct advantage of \textsc{Nupunkt} is that per-book unsupervised learning leads
to improvements in segmentation quality.

A different alternative would be to use \textsc{SaT} on more languages. Annotated training data can
be used to fine-tune \textsc{SaT} models to particular languages. In an ideal scenario,
we would use high-quality training data in low-resource languages to generate more
reliable outputs.

\paragraph{On the Lack of SaT Batching.}

The \textsc{SaT} library allows batching, but we do not batch the books. This is because the
typical book text is long, and internally \textsc{SaT} splits long inputs into overlapping
$512$ token windows and batches those tokens through the model. A single medium-length book can
saturate a GPU on its own. Small batch testing suggests that batching books into groups of $200$
(and allowing \textsc{SaT} to generate appropriate batches internally) improves throughput
between $15$ and $25$ percent. This is a notable increase, but does not justify the added complexity
for only \numprint{20629} books.
Instead, the pipeline is organized for simple error tracking and resuming after faults.

\subsection{Alternate Chunking Algorithms Considered}\label{appendix:ss:chunking}

Many high-quality chunking algorithms require a GPU in practice.
For example, one can use a BERT or RoBERTa model fine-tuned on training data.
We did not find any GPU-based chunking algorithm that fit within our compute constraints using
available multilingual training data.

A priori, we expected a small variant of the sentence segmentation algorithms to be serviceable.
Some sentence segmentation algorithms assign probabilities for tokens to be the end of a sentence.
We hypothesized that choosing a high threshold for this probability could break chunks of sentences
into paragraphs, but experiments showed that this approach did not work well in practice.

We found TextTiling~\parencite{hearst1997text} and C99~\parencite{choi2000advances} to be computationally
viable. When combined with modern multilingual sentence embeddings, they are both performant and
customizable. The pipeline code repository contains complete modern implementations of both a
TextTiling-based algorithm and a C99-based algorithm. We ultimately chose TextTiling because it led
to better duplicate detection later in the pipeline.

\subsection{Duplicate Identification Details}

\paragraph{Band Bucket Sizes.}

Each 128-bit simhash is split into 6 bands of (22, 21, 21, 21, 21, 22) bits. We consider two hashes
with Hamming distance at most $5$ to be duplicates. Any such pair must have at least one identical
band, and hence to identify candidate duplicate pairs we search for exact matches across each of the
$6$ bands.

Each band has at least $21$ bits and there are $\approx 1.4 \times 10^9$ paragraphs. Each band
holds on average
\begin{equation*}
  \frac{\#\, \textup{paragraphs}}{\# \textup{possible values}}
  \approx
  \frac{1.4 \cdot 10^9}{2^{21}}
  \approx
  667
\end{equation*}
different paragraphs. Thus after restricting to buckets formed from exact band equality, buckets are
small enough on average to allow exhaustive comparison (each requiring two \texttt{xor}s, two
\texttt{popcount}s, and one addition over 64-bit limbs). We implemented an upper limit: if a bucket
has more than 30k items, it logs a warning and does not compare within the bucket.
This limit is never reached for this collection.

\paragraph{Resource Footprint.}

The global simhash array (\texttt{hashes.bin}, $\approx$23GB) is memory-mapped and
available to individual worker processes. The union-find data structure is managed by a parent
process and individual workers require almost no private state in addition to access to the page-cached
simhash array. Each worker computes Hamming distances in parallel and streams confirmed pairs to the
parent process. Confirmed near-duplicate pairs are merged with a union-find data structure keyed on
the transient flat paragraph index. This structure stores up to one index per paragraph, as well as
up to one byte to track rank for each paragraph (1.4B entries $\times$ (8 byte index $+$ 1 byte
rank), roughly 13GB in total). The duplicate identification pipeline fits in 48GB of RAM.

\subsection{Bits-Per-Byte Details}\label{appendix:ss:bpb}

The bits-per-byte computation is a thin wrapper around direct calls on the tokenizer and model
from \textsc{Qwen/Qwen3-0.6B-Base} using the \textsc{transformers} library and
\textsc{PyTorch}~\parencite{qwen3technicalreport, wolf-etal-2020-transformers, pytorch}.
The implementation closely follows standard \textsc{PyTorch} and \textsc{transformers} usage, except
that \textsc{Qwen3} models require right padding rather than left padding during batching.

\paragraph{Determining the Language Model.}

We initially sought an inexpensive but reliable proxy for perplexity (before switching to
per-byte normalization for BPB instead of per-token normalization).
We investigated whether easily computable features could accurately predict perplexity.

To make this concrete, we studied whether these features allow accurate prediction of \textsc{Qwen3}
perplexity across models of different sizes ranging from 0.6B parameters to 14B parameters.
We trained simple Ridge linear regression models on an 80-20 train-test split from $3067$ random
pages of text from IB-HL to explore feasibility.

\begin{enumerate}[nosep,noitemsep]
  \item N-gram complexity from a complete-but-efficient n-gram language model
    like \\
    KenLM~\parencite{kenlm} in a few variants:
    \begin{itemize}[nosep,noitemsep]
      \item character n-grams,
      \item word n-grams based on
        SentencePiece~\parencite{sentencepiece} subword \\ units~\parencite{sennrich-etal-2016-neural}, or
      \item \textsc{Qwen3} tokenizer n-grams (where n-grams consist of sets of $n$ tokens coming from
        a \textsc{Qwen3} tokenizer).
    \end{itemize}
  \item Embeddings, predictions, and confidences from the Endmatter classifier (cf.\
    \S\ref{ssec:endmatter}),
  \item Fundamental natural language processing descriptors, including
    \begin{itemize}[nosep,noitemsep]
      \item character count,
      \item word count,
      \item ratio of characters to newlines,
      \item ratio of characters to whitespace,
      \item ratio of characters that are digits,
      \item number of unique chars,
      \item number of bigrams,
      \item bigram entropy, and
      \item rare bigram ratios.
    \end{itemize}
\end{enumerate}

\begin{figure}[htb!]
  \centering
  \includegraphics[width=\textwidth]{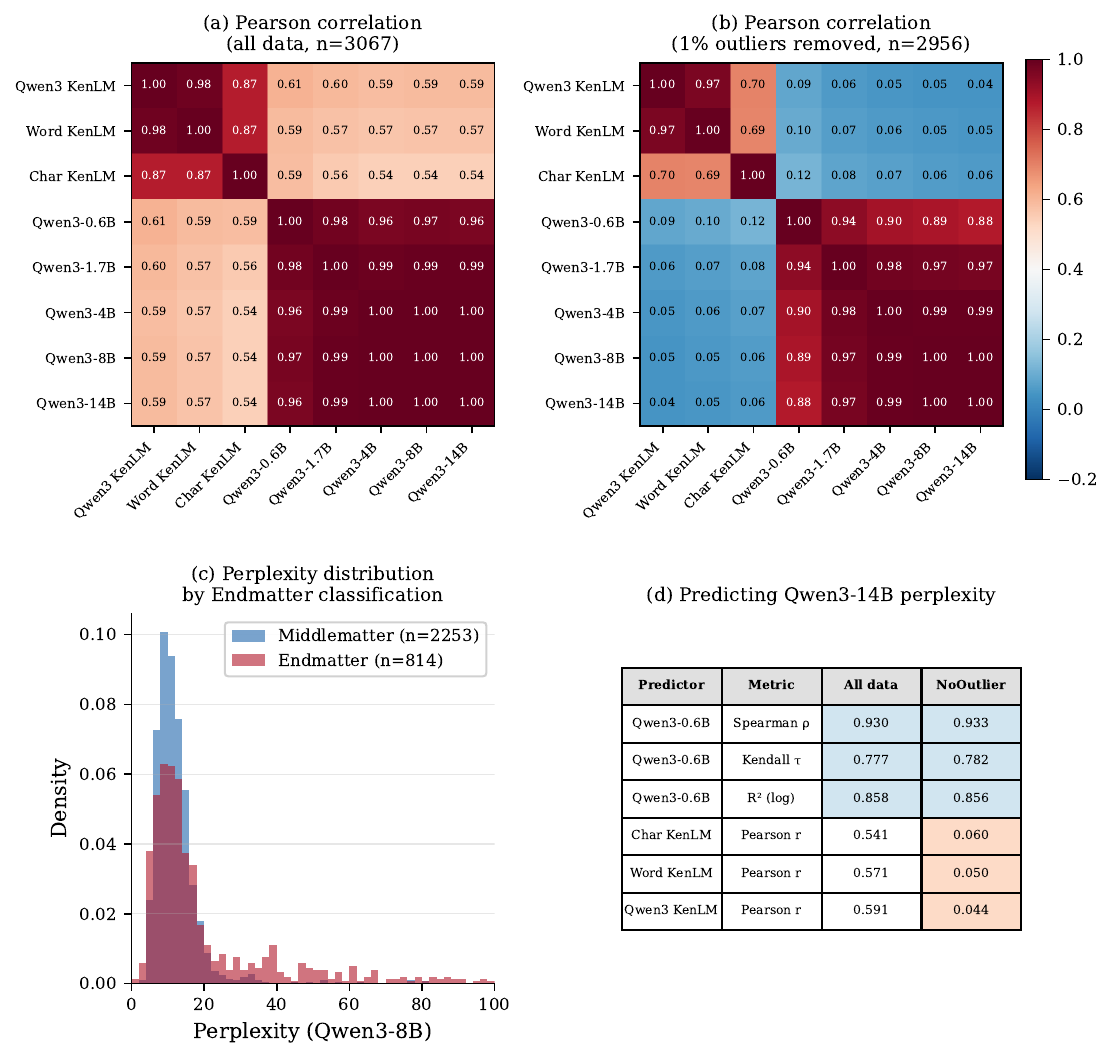} 
  \caption{Statistical correlations of \textsc{Qwen3} perplexity with other measures}%
  \label{fig:perplexity_comparison}
\end{figure}

See Figure~\ref{fig:perplexity_comparison} for correlations and prediction behavior.
We observed that multiple features carried predictive power for extremely high perplexity
(especially the rare bigram ratio and all KenLM language models). After removing
the 1\% outliers,\footnote{We removed the top and bottom 1\% for each of the
  three ngram metrics for panel (b). Many pages were simultaneous outliers,
leading to the removal of only 111 pages.}
we observed almost no correlation between actual perplexity and estimated perplexity (panel (b)).
The bimodal aspect of endmatter classification leads to poor predictive behavior (though
\emph{extreme} perplexity does weakly correlate with being endmatter).

We did not anticipate the difference in behavior between panel (a) and panel (b) in
Figure~\ref{fig:perplexity_comparison}.
The predictions from n-gram language models correlate strongly with each other (even if they use
different atomic units of text), but do not correlate with actual perplexity.
One should expect \textsc{Qwen3} models with similar parameter counts to have higher correlation,
but it was not obvious that the smallest model, 0.6B, was highly predictive of the 14B model.
Further, panel (d) shows that the relative ordering between perplexities is largely
preserved (Kendall $\tau = 0.782$).
From this, we determined that we would use the smallest, most computationally efficient model.


\paragraph{Error Rates in Computing BPB.}

After running the full pipeline, we noted that a different implementation of computing
cross-entropy would allow effective computation of BPB values even for long paragraphs.
Specifically, one could implement \emph{chunked} or \emph{fused}
cross-entropy~\parencite{hsu2024liger}.
Alternatively, one could use multiple GPUs together instead of many independent GPUs.
We did not revisit the BPB values of the 4 too-short or the \numprint{12391} too-long paragraphs.

\section{Examples with Known Limitations}\label{appendix:s:bad}

Though we have tried to be conservative in cleaning and applying changes, we know of examples where
the current pipeline has unintended side effects. We give two examples here.

\begin{figure}[h!]
  \centering
  \includegraphics[width=4in]{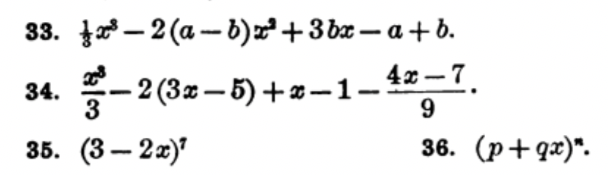}

  \vspace*{1em}

  \begin{minipage}{0.45\textwidth}
  \begin{minted}[frame=single]{text}
33. 2-2 (a - b) x²+3 bx — a+b.
-
34. 1 — 2(3x−5) + x − 1 −
3
35. (3-2x)
−
4x-7
9\end{minted}
  \vspace*{-2em}
  \begin{center}{Source OCR Text}\end{center}
  \end{minipage}
  \hspace{1em}
  \begin{minipage}{0.45\textwidth}
  \begin{minted}[frame=single]{text}
33. 2-2 (a - b) x²+3 bx — a+b.
34. 1 — 2(3x-5) + x - 1 3
35. (3-2x)
-4x-7
9\end{minted}
  \vspace*{-2em}
  \begin{center}{Dehyphenated Text}\end{center}
  \end{minipage}

  \caption{Example of Meaning-Altering Dehyphenation Taken from \textsc{32044000046607}.}\label{figure:badmath}
\end{figure}

\paragraph{Dehyphenation and Mathematics.}
The current pipeline does not treat math differently from prose (nor does the currently available
OCR). In technical texts, minus signs and fraction bars might be treated as end-of-line hyphens and
potentially removed during dehyphenation (\S\ref{ssec:dehyphenation}).
For example, Figure~\ref{figure:badmath} is taken from \emph{A first course in the differential and
integral calculus} (barcode \textsc{32044000046607}), page 448 (with page number 426 in the book).

Figure~\ref{figure:badmath} shows three lines of exercises from a book, the source OCR output
(including original newlines), and the output after dehyphenation. Three changes in
hyphenation occur in this example:
\begin{enumerate}
  \item The lone hyphen on the second line of the source OCR is removed.
  \item The hyphen at the end of the exercise 34 line in the OCR text is removed, causing \texttt{x - 1
    - \textbackslash n 3} to be replaced by \texttt{x - 1 3}.
  \item The lone hyphen on the line after exercise 35 in the source OCR is kept, but the newline is
    removed and now the \texttt{4x-7} on the following line becomes \texttt{-4x-7}.
\end{enumerate}
All three affect the meaning of the source OCR. But the source OCR does not capture the full
meaning of the original math exercises.\footnote{
  One might ask where the three hyphens came from. The first likely originated from the
  fraction bar in either $\tfrac{1}{3}$ from exercise 33 or $\tfrac{x^2}{3}$ from exercise 34; the
  second has to do with the OCR engine determining that $\frac{4x-7}{9}$ belongs in a column with
  exercise 36, so ordering is confusing; and the third is probably the misplaced fraction bar from
  $\frac{4x-7}{9}$.}

These cases are rare relative to the billions of correct merges. But these examples suggest that
technical and mathematical volumes are disproportionately affected.

\begin{figure}[h!]
  \centering
  \fbox{\includegraphics[width=5in]{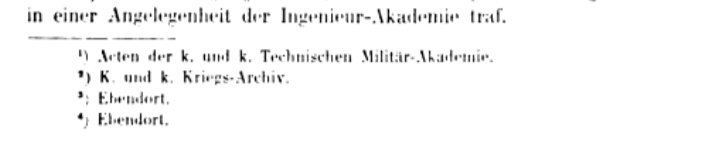}}

  \fbox{\includegraphics[width=5in]{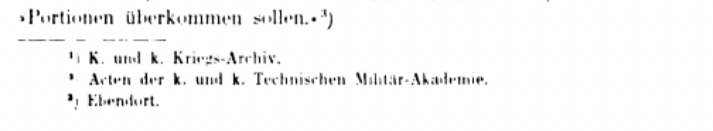}}

  \fbox{\includegraphics[width=5in]{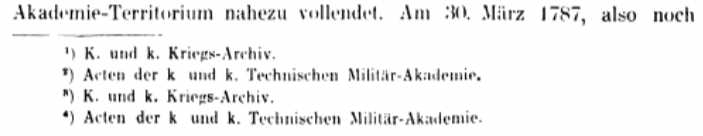}}

  \caption{Example of Removing Citations in Footers Taken from \textsc{32044004475331}.}\label{figure:badfooter}
\end{figure}

\paragraph{Removing Citation Footers.}
Header/footer removal in this pipeline targets repetitive lines near the top or bottom of many
nearby pages. Some books use abbreviated citation formats in footers or marginalia. When many
nearly-identical citations occur in close proximity, they may all be removed.
Figure~\ref{figure:badfooter} shows three footers of pages in \emph{Geschichte der K.\ und K.\
Technischen Militär-Akademie} (barcode \textsc{32044004475331}).

These footnotes include repeated references to \texttt{K.\ und k.\ Kriegs-Archiv.},
and \texttt{Ebendort.}. Similar citation patterns occur throughout the book. As these lines recur at
the foot of the page, the remover treats them as running footers. The number of citations to
\texttt{Kriegs-Archiv} decreases from $137$ occurrences in the source OCR to $64$ in the output.

Currently all marginalia are OCRed and grouped with the main text, causing remaining bibliographic
references to be placed in the middle of paragraphs. Current treatment of marginalia at both the OCR
level and the cleaning level does not allow accurate tracking of bibliographic provenance. We regard
restoring these paratextual elements as a worthwhile goal for future work.

\section{Using the Parser Library}\label{appendix:ibet_parser}

\begin{sloppypar}
We wrote a lightweight, pure-Python, dependency-free parser
library\footnote{\url{https://github.com/institutional/institutional-books-enriched-text-parser}}
and tool that can be used to filter and iterate through the final annotated dataset.
The library uses \texttt{html.parser.HTMLParser} from the Python standard library as the core HTML
parser.
\end{sloppypar}

\begin{table}[tbh!]
\centering
\footnotesize
\setlength{\tabcolsep}{4pt}
\renewcommand{\arraystretch}{1.15}
\begin{tabular}{@{}l l p{5.5cm}}
\toprule
\textbf{Object field} & \textbf{Type} & \textbf{Description} \\
\midrule
\texttt{paragraph.text} & string & Content of the paragraph \\
\texttt{paragraph.bpb} & float & \textsc{Qwen/Qwen3-0.6B-Base} computed bits-per-byte \\
\texttt{paragraph.language} & string & ISO 639-3 code of language detected for the paragraph \\
\texttt{paragraph.is\_duplicate} & boolean & Whether this paragraph is a duplicate of another
paragraph in the whole collection. \\
\texttt{section.paragraphs} & list[paragraph] & Paragraph objects contained in section \\
\texttt{section.bpb} & float & Average bits-per-byte of contained paragraphs \\
\texttt{book.bpb.\{p10,p30,median,p70,p90,avg\}} & float & BPB percentile and average values \\
\bottomrule
\end{tabular}
  \captionsetup{width=0.8\linewidth}
  \caption{IBET Parser Object Description}\label{table:ibet_parser}
\end{table}

The library exposes its functionality through the \texttt{BookDataset} class, which takes any
iterable of Python dictionaries with a schema matching the dataset fields described in
Appendix~\ref{appendix:s:dataset_fields}. This can take many forms, such as a list of manually
curated dictionaries, partial outputs from the pipeline, or the streamed HuggingFace
dataset~\parencite{datasets} produced by this pipeline. For example:

\newpage{}

\begin{minted}[frame=single, framesep=2mm]{python}
from datasets import load_dataset
from ibet_parser import BookDataset

ds = load_dataset(
    "institutional/institutional-books-hl-enriched-text",
    split="train", streaming=True
)
books = BookDataset(ds)
\end{minted}

Once \texttt{BookDataset} is instantiated, one can filter and iterate through books, sections, and
paragraphs:
\begin{minted}[frame=single, framesep=2mm]{python}
from itertools import islice

for book in islice(books, 10):     # for demonstration, take 10 books
    print(book.barcode, book.primary_language, book.token_count)
    for paragraph in book.paragraphs:
        print(paragraph.text[:100])
        # do_something_with_paragraph(paragraph)

for book in islice(books, 10):
    for section in book.sections:
        # do_something_with_section(section)
        for paragraph in section.paragraphs:
            print(paragraph.text[:100])
        print()                    # add newline between sections
\end{minted}
Books at the dataset level can be filtered by language or token count range.
Specifically, \mintinline{python}{books.filter()} accepts optional \texttt{language},
\texttt{token\_count\_min}, or \texttt{token\_count\_max} arguments. Both token counts accept
integer arguments and restrict to books with the specified token range; \texttt{language} accepts
either one or a list of ISO 639-3 language strings and restricts to books whose primary language is
one of those specified.
For example, to iterate over English books with at least 1000 tokens:
\begin{minted}[frame=single, framesep=2mm]{python}
for book in books.filter(language='eng', token_count_min=1000):
    # do_something_with_book(book)
    pass
\end{minted}

For paragraph-level filtering: \mintinline{python}{books.paragraphs.filter()} accepts
\texttt{language} (as a single string or a list, as above), \texttt{deduplicated} (Boolean: if true,
exclude duplicate paragraphs from iteration), \texttt{bpb\_min} and \texttt{bpb\_max} (as floats).
The per-volume BPB values are available as attributes for convenient reference (cf.\
Table~\ref{table:ibet_parser}).
Section filtering uses the same semantics on \mintinline{python}{books.sections.filter()} and
passes each filter to the paragraphs within each section. Sections where all paragraphs are filtered
out are entirely excluded from iteration.

\end{appendices}

\end{document}